\documentclass{article}
\usepackage{iclr2026_conference,times}

\usepackage{amsmath,amsfonts,bm}

\def\eqref#1{equation~\ref{#1}}

\def\1{\bm{1}}

\DeclareMathAlphabet{\mathsfit}{\encodingdefault}{\sfdefault}{m}{sl}
\SetMathAlphabet{\mathsfit}{bold}{\encodingdefault}{\sfdefault}{bx}{n}

\usepackage{hyperref}
\usepackage{url}
\usepackage{amsmath,amssymb,amsfonts,bm}
\usepackage{graphicx}
\usepackage{xcolor}
\usepackage{booktabs}
\usepackage{multirow}
\usepackage{array}
\usepackage{colortbl}
\usepackage{tabularx}
\usepackage{makecell}
\usepackage{wrapfig}
\usepackage{subcaption}
\usepackage{float}
\usepackage{enumitem}
\usepackage{pgfplots}
\pgfplotsset{compat=1.18}
\usepgfplotslibrary{groupplots}

\usepackage[most]{tcolorbox}
\tcbuselibrary{breakable, skins, theorems}

\newtcolorbox{hypothesisbox}[2][]{%
    enhanced, breakable,
    colback=gray!4!white, colframe=violet!55!black, coltitle=white,
    fonttitle=\bfseries\scshape, title={#2},
    arc=2pt, outer arc=2pt, boxrule=0.8pt,
    left=10pt, right=10pt, top=8pt, bottom=8pt,
    attach boxed title to top center={yshift=-2.5mm,yshifttext=-1mm},
    boxed title style={colback=violet!55!black, colframe=violet!55!black, arc=2pt, boxrule=0pt, left=8pt,right=8pt},
    #1
}

\newtcolorbox{findingbox}[2][]{%
    enhanced, breakable,
    colback=teal!3!white, colframe=teal!60!black, coltitle=white,
    fonttitle=\bfseries\scshape, title={#2},
    arc=2pt, outer arc=2pt, boxrule=0.7pt,
    left=10pt, right=10pt, top=6pt, bottom=6pt,
    attach boxed title to top center={yshift=-2.5mm,yshifttext=-1mm},
    boxed title style={colback=teal!60!black, colframe=teal!60!black, arc=2pt, boxrule=0pt, left=8pt, right=8pt},
    #1
}

\definecolor{groundedTeal}{HTML}{009E73}     
\definecolor{priorSlate}{HTML}{6E747C}       
\definecolor{invertedVerm}{HTML}{D55E00}     
\definecolor{neutralGray}{HTML}{BFC4CB}      
\definecolor{accentBlue}{HTML}{0072B2}       
\definecolor{groundedTint}{HTML}{D6F0E5}
\definecolor{priorTint}{HTML}{E5E7EA}
\definecolor{invertedTint}{HTML}{F8DECD}
\definecolor{accentTint}{HTML}{D6E7F3}
\definecolor{highlightTint}{HTML}{FFF6B8}    

\colorlet{rowcatA}{invertedTint}
\colorlet{rowcatB}{priorTint}
\colorlet{rowcatC}{accentTint}
\colorlet{rowcatD}{invertedTint}
\colorlet{rowcatE}{groundedTint}
\colorlet{rowours}{highlightTint}
\colorlet{markgreen}{groundedTeal}
\colorlet{markred}{invertedVerm}
\colorlet{markamber}{accentBlue}
\usepackage{pifont}
\usepackage{rotating}

\newcommand{\cmark}{\textcolor{groundedTeal}{\ding{51}}}
\newcommand{\xmark}{\textcolor{neutralGray!85!black}{\ding{55}}}
\newcommand{\pmark}{\textcolor{accentBlue}{\ding{108}}}
\newcommand{\rothdr}[1]{\rotatebox{60}{\makecell[l]{\scriptsize #1}}}
\newcommand{\subhead}[1]{\smallskip\noindent\textbf{#1}}

\setcitestyle{numbers,square,sort&compress,citesep={,}}

\title{Decodable Is Not Grounded: \\
A Vision-Ablation Arbiter for VLM Spatial Reasoning}

\author{
Chih-Ting Liao\thanks{Project lead. Corresponding author: \texttt{mill.liao@unsw.edu.au}} \\
University of New South Wales \\
\texttt{mill.liao@unsw.edu.au}
\And
Fei Shen\thanks{Corresponding author: \texttt{shenfei29@nus.edu.sg}} \\
National University of Singapore \\
\texttt{shenfei29@nus.edu.sg}
\And
Xin Cao \\
University of New South Wales
\And
Tat-Seng Chua \\
National University of Singapore
}

\iclrfinalcopy

\begin{document}

\maketitle
\lhead{Preprint}\rhead{}

\begin{abstract}
The standard way to read latent knowledge out of a model, a linear probe confirmed by a steering
recovery, can systematically \emph{overstate} what a vision--language model (VLM) actually grounds in
the image. We show this on spatial reasoning, where the error is invisible to both probing and steering
yet exposed by a one-line causal control: replacing the image with a gray blank. Probes decode the
within-axis answer at $73$--$97\%$ across axes, and a training-free projection lifts a near-chance axis
from $59\%$ to $79\%$, exactly the signature of unlocking latent knowledge. The blank-image arbiter
refutes it, revealing three grounding regimes that probing conflates: an axis can be \emph{grounded}
(vision-dependent, correct), a \emph{prior} (vision-independent, with its decode and its apparent
recovery a directional default rather than perception), or, surprisingly, \emph{inverted}: decodable,
causally controllable, but deployed with the wrong sign, so the model scores \emph{below} chance and the
error \emph{requires} looking. The taxonomy holds across the studied VLMs: in fourteen models spanning
six language-model families and $2$B--$27$B, horizontal is grounded, vertical is a prior, and depth is
inverted, with the inversion emerging at scale within families. The decode$\neq$deploy inversion
replicates on seven of eight models across five families, and the minimal edit that re-deploys it varies
with geometry: a training-free rotation matches a trained edit on the cleanest model, while distributed
inversions need a trained low-rank edit, tracing a per-model correction-complexity spectrum. The cheap,
self-calibrating arbiter cleanly separates grounded perception, inverted perception, and prior
substitution; we argue it should be a default control for latent-knowledge and steering claims in VLMs.
\end{abstract}

\section{Introduction}
\label{sec:intro}

A vision--language model (VLM) can answer a spatial question \emph{below} chance, and blinding it
(replacing the image with a same-size gray blank) makes it \emph{more} accurate: the model is not
failing to use the image, it is using it backwards. This sits inside a more familiar pattern. A linear
probe reads a target attribute out of activations even when the model answers questions about it
poorly~\citep{gurnee2024space, linearspatiotemporal2026, seeingbutnotbelieving2025}, and the standard
reading is that the model \emph{represents} the attribute but fails to \emph{deploy} it, motivating
training-free interventions (steering, projection, representation engineering) that unlock the latent
capability without gradient updates~\citep{rimsky2024caa, zou2023representation,
linearspatiotemporal2026}. This paper is a cautionary study of that inference: we follow the recipe to
its natural conclusion on one model, Qwen2.5-VL-7B~\citep{qwen2vl}, and one controlled axis (egocentric
directional judgement on ViewSpatial-Bench~\citep{viewspatial2025}), then show the conclusion is wrong
and that the corrected picture holds across fourteen models and six LM families (\S\ref{sec:crossmodel}).

\subhead{The seductive result.} Camera-relative direction decomposes into horizontal (left/right),
vertical (above/below), and depth (front/back). The model is strong on horizontal, near chance on
vertical, and \emph{below} chance on depth (\S\ref{sec:dissociation}), yet a linear probe decodes all
three axes from the residual stream, vertical the \emph{most} ($94\%$ vs.\ horizontal's $85\%$). A
training-free projection along the vertical direction then lifts vertical accuracy from $59\%$ to
$79\%$ while leaving horizontal untouched (\S\ref{sec:intervention}), precisely the result one would
publish as latent spatial knowledge recovered without training.

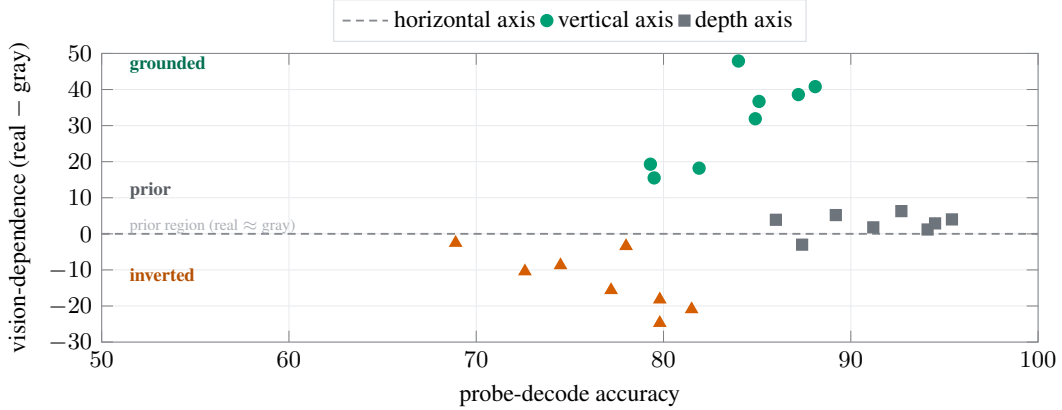
\begin{figure}[t]
\centering
\begin{tikzpicture}
\begin{axis}[
    width=\linewidth, height=5.4cm,
    xlabel={\small probe-decode accuracy},
    ylabel={\small vision-dependence (real $-$ gray)},
    xmin=50, xmax=100, ymin=-30, ymax=50,
    xtick={50,60,70,80,90,100},
    ytick={-30,-20,-10,0,10,20,30,40,50},
    tick label style={font=\footnotesize},
    grid=major, grid style={neutralGray!35, very thin},
    axis line style={neutralGray!70!black},
    legend style={at={(0.5,1.04)}, anchor=south, legend columns=3,
                  font=\footnotesize, draw=neutralGray!60},
    legend cell align=left,
    every axis plot/.append style={line width=0.6pt},
]
\addplot[neutralGray!70!black, densely dashed, no marks, line width=0.7pt]
    coordinates {(50,0) (100,0)};
\node[anchor=west, font=\tiny, neutralGray!90!black]
    at (axis cs:51,2) {prior region (real $\approx$ gray)};

\addplot[only marks, mark=*, mark size=2.2pt, groundedTeal] coordinates {
    (84.9,31.9)   
    (87.2,38.6)   
    (84,47.9)   
    (79.3,19.3)   
    (88.1,40.8)   
    (85.1,36.7)   
    (79.5,15.5)   
    (81.9,18.2)   
};
\addlegendentry{horizontal axis}

\addplot[only marks, mark=square*, mark size=2.0pt, priorSlate] coordinates {
    (94.1,1.2)   
    (95.4,4)   
    (91.2,1.8)   
    (87.4,-3)  
    (92.7,6.3)   
    (94.5,2.9)   
    (86,3.9)   
    (89.2,5.2)   
};
\addlegendentry{vertical axis}

\addplot[only marks, mark=triangle*, mark size=2.4pt, invertedVerm] coordinates {
    (77.2,-15.6)  
    (79.8,-24.7)  
    (72.6,-10.4)  
    (68.9,-2.5)  
    (79.8,-18.2)  
    (81.5,-20.9)  
    (74.5,-8.7)  
    (78,-3.4)  
};
\addlegendentry{depth axis}

\node[anchor=south west, font=\scriptsize\bfseries, groundedTeal!70!black]
    at (axis cs:51,42) {grounded};
\node[anchor=south west, font=\scriptsize\bfseries, priorSlate!80!black]
    at (axis cs:51,7) {prior};
\node[anchor=north west, font=\scriptsize\bfseries, invertedVerm!85!black]
    at (axis cs:51,-7) {inverted};
\end{axis}
\end{tikzpicture}
\caption{\textbf{Decodability does not predict whether behavior depends on the image.} Each point is a
(model, axis) pair across 8 VLMs: $x$ is probe decoding accuracy, $y$ is \emph{vision-dependence}
(real-image $-$ gray-image behavior). Probes decode every axis well ($x{>}70\%$), but behavior splits
into three regimes probing cannot see: horizontal grounded (high), vertical a prior (zero), depth
inverted (below zero). The arbiter recovers this in one forward pass. Percentages.}
\label{fig:thesis}
\end{figure}

\subhead{The arbiter.} We then ask what the probe-and-steer pipeline never asks: does the recovered
behavior \emph{depend on the image}? A one-line control (replace the image with a same-size gray blank,
leaving question and options unchanged; Fig.~\ref{fig:thesis}) fractures the single not-deployed story
into three grounding regimes. \textbf{Horizontal} is grounded and correct (accuracy collapses
$82\%\to49\%$ when blinded). \textbf{Depth} is grounded and \emph{inverted} (below chance with the
image, $31\%$; reverts to chance without it, $45\%$, so the error \emph{requires} looking).
\textbf{Vertical} is \emph{not grounded}: real and blank accuracy are identical ($59\%\approx60\%$), and
the projection recovery gains the same with no image ($+19$ blind vs.\ $+21$ sighted), so the $94\%$
decode reads a directional default-to-above prior, not vision.

\subhead{Why this matters.} Vertical is the cautionary case: the \emph{most} decodable axis, the
cleanest training-free recovery, and the one axis that does not use vision, exactly the headline a
decode-and-steer pipeline would report. Depth is the surprise: a model that scores \emph{below} chance
because it reads a real visual cue with the wrong sign. The vision-ablation arbiter catches both, is
essentially free (one forward pass with pixels zeroed), and is \emph{discriminating} rather than
deflationary. \textbf{Our contributions:} \textbf{(i)} a three-regime grounding taxonomy
(grounded-correct, grounded-inverted, prior-driven) and the finding that linear decodability and
behavioral accuracy each fail to identify which regime an axis is in, the most decodable and most
recoverable axis being the least grounded (\S\ref{sec:dissociation}--\ref{sec:arbiter});
\textbf{(ii)} a cheap, discriminating vision-ablation arbiter, robust across five ablations including a
real-but-mismatched in-distribution image, which we argue should be a default control for
latent-knowledge and steering claims in VLMs (\S\ref{sec:arbiter}, \S\ref{sec:discussion});
\textbf{(iii)} architecture-generality and a task-type boundary across fourteen VLMs, six LM families,
and $2$B--$27$B: across capable models horizontal is grounded, vertical a prior, depth inverted
(emerging with scale within families), and the same models that invert camera-egocentric depth are
grounded-correct on near-field and metric depth (\S\ref{sec:crossmodel}--\ref{sec:external}); and
\textbf{(iv)} a correction that reads out the geometry: the inversion replicates on seven of eight
models across five families, and the minimal re-deploying edit (training-free rotation on the cleanest
model, matching a trained edit; a trained low-rank edit on distributed ones) traces a per-model
correction-complexity spectrum (\S\ref{sec:method}).

\section{Related Work}
\label{sec:related}

\subhead{Spatial reasoning in VLMs.} Early benchmarks test two-object relations from canonical
viewpoints (What'sUp \citep{kamath2023whatsup}, VSR \citep{liu2023vsr}), now largely saturated
\citep{cai2024spatialbot, du2024embspatial}; the frontier targets egocentric and multi-perspective
settings: VSI-Bench \citep{yang2024thinking}, MMSI-Bench \citep{yang2025mmsi}, OmniSpatial
\citep{omnispatial2026}, and ViewSpatial-Bench \citep{viewspatial2025}, the source of our task. A large
training-based literature closes these gaps with synthetic 3D supervision or RL
\citep{chen2024spatialvlm, spatialssrl2025, daxberger2025mmspatial}; our study is diagnostic and
training-free, orthogonal to those remedies.

\subhead{Linear probing, spatial IDs, and probe validity.} Linear world models for space have
been read out of LLMs \citep{gurnee2024space, linearspatialworld2025} and VLMs; most relevantly,
\citet{linearspatiotemporal2026} identify linear \emph{spatial IDs} that are causally steerable across
VLMs. This establishes that spatial structure is linearly tractable and steerable; we ask the
complementary question, whether a decodable, steerable direction is the one the model \emph{uses to
look}, and find the most decodable axis is the one that ignores the image. This connects to a
long-standing caution: high decoding accuracy certifies linear presence, not use, and control tasks are
needed to calibrate it \citep{hewitt2019control, belinkov2022probing}. We provide a \emph{behavioral,
causal} control (vision ablation) rather than a probe-side one.

\subhead{Latent spatial knowledge: present, but used how?} A fast-growing line argues VLMs \emph{do}
represent space and that failures are an access problem. Closest to us, \citet{scenetopology2026} show
a latent 3D-topology map exists but is overshadowed by semantics and \emph{surface} it with a
subspace + Dirichlet regularizer; \citet{beyondsemantics2025} blame vision-token norms suppressing
positional cues and restore it by \emph{normalizing} norms; \citet{geomfailures2026} explain
multi-object errors as concept-vector \emph{interference} and validate by steering; and concurrently
\citet{whyfarup2026} find a \emph{vertical--distance entanglement} that strengthens with scale. These
share a template (locate a representation that is absent, overshadowed, entangled, or steerable, then
surface or steer it) and assume the obstacle is \emph{access}. Our phenomenon differs in kind: depth is
present, correctly decodable, and causally controllable yet \emph{deployed with the wrong sign}
(decode$\neq$deploy), while vertical is decodable but behaviorally a prior. Our discriminator is a
\emph{behavioral} vision-ablation arbiter, and our remedy is a measured \emph{correction-complexity
spectrum}. A capability matrix (Appendix~\ref{app:additional}, Table~\ref{tab:related}) places these works on common axes; the inversion regime and the
cross-model correction ladder are, to our knowledge, unique here.

\subhead{Steering and the actionability gap.} Contrastive activation addition \citep{rimsky2024caa,
panickssery2024} and representation engineering \citep{zou2023representation}, following the
activation-editing tradition \citep{meng2022rome, wang2023interpretability}, inject a direction to
change behavior. \citet{basu2026interp} give a pointed negative result: probes reach near-perfect AUROC
while interventions correct only $\sim$20\% of errors. We sharpen this from another angle: an
intervention that \emph{moves} accuracy can still act on a non-visual prior, so a behavioral success
is not grounded recovery without a causal control. Relatedly, the VQA literature showed
models exploit question priors and score well without the image \citep{goyal2017vqa,
agrawal2018dontassume}. Our arbiter shares that intuition but is applied \emph{per axis} and reads the
\emph{signed} real$-$ablated difference, so it separates not just grounded from ungrounded but a regime
a whole-question blind baseline cannot express: an \emph{inverted} axis, image-driven yet deployed
wrong-sign (below chance with the image, at chance without). \citet{seeingbutnotbelieving2025} document
a related attention--correctness disconnect on spatial QA.

\section{Setup}
\label{sec:setup}

\subhead{Task and axes.} Our probe is one controlled task that cleanly separates three spatial axes.
We use the Camera-Relative-Direction split of ViewSpatial-Bench \citep{viewspatial2025}: egocentric
four-option questions ``where is $X$ relative to $Y$ from the camera's perspective'' ($1773$ items).
We decompose each item by the physical axis its gold answer lies on (\emph{horizontal} left/right,
\emph{vertical} above/below, \emph{depth} front/back, and \emph{composite}, diagnostic only) and
run all causal experiments on the $1047$ pure-axis items (per-axis counts in
Table~\ref{tab:dissociation}). We report native \textbf{18-way} accuracy (chance $25\%$) and a \textbf{binary} within-axis
accuracy (the two poles, order randomized; chance $50\%$), balanced wherever a pole split is skewed.
The anchor is Qwen2.5-VL-7B-Instruct \citep{qwen2vl}; implementation details are in
Appendix~\ref{app:additional}.

\subhead{Probing, steering, and the arbiter.} For decodability we fit a five-fold CV logistic probe
(PCA-50) on the decision-token residual stream per axis and layer, reporting accuracy over a
shuffled-label floor. For causal tests we extract a per-axis direction (diff-of-means or back-projected
probe weight) and add it at the decision token scaled in residual-stream standard deviations; the
\emph{training-free projection} amplifies the localized direction over the top-$k$ neurons in the
L20--24 band with a single global scalar $\gamma$ and no gradients or labels (details in
Appendix~\ref{app:additional}). The \textbf{vision-ablation arbiter} re-evaluates any condition with
the image replaced by a same-size gray blank, prompt and decoding unchanged, and classifies each axis
by the signed real$-$gray gap: \emph{grounded} if real exceeds gray and gray falls to chance,
\emph{inverted} if real is below chance and gray reverts toward it, and \emph{prior} if real and gray
coincide within bootstrap CIs. We apply it to both baseline
accuracy and the projection gain, and guard against input-degeneracy with five ablations (gray, black,
noise, patch-scramble, and a real-but-mismatched same-axis image) required to agree
(\S\ref{sec:arbiter}). Intervention hyperparameters ($\gamma$, $\theta$, layer band, top-$k$) are
chosen on validation and frozen for the reported test numbers.

\subhead{Generality battery.} The deep mechanism is characterized on the anchor; for generality we
run an independent battery (per-axis behavior, arbiter, per-layer probe) on thirteen further VLMs
spanning six LM families and 2B--27B (full roster in \S\ref{sec:crossmodel}), with per-model
decode-peak layers re-found. We further run the behavioral battery on two external benchmarks
(What'sUp \citep{kamath2023whatsup}, 3DSRBench \citep{ma20243dsrbench}) and a nine-method depth
correction battery (\S\ref{sec:method}).

\section{The Decode\texorpdfstring{$\neq$}{!=}Deploy Dissociation and a Vision-Ablation Arbiter}
\label{sec:dissociation}

\subhead{Behavior: a single grounded axis.} The native task accuracy is $45.4\%$ on the full
$1773$ items, matching prior ViewSpatial numbers and validating our evaluation. Decomposed by axis
(Table~\ref{tab:dissociation}), the model is strong on horizontal ($74.9\%$ 18-way, $81.3\%$ binary),
near chance on vertical ($26.9\%$ / $58.5\%$), and \emph{below} chance on depth ($16.3\%$ / $30.7\%$;
binary depth is below chance at $p<10^{-5}$, binomial).
Controls rule out shortcuts (a same-dimension distractor tracks full accuracy; the composite class
scores above chance \emph{only} with a left/right component, $52.1\%$ vs.\ $9.2\%$;
Appendix~\ref{app:robustness}): the model is competent on horizontal and defaults elsewhere. Depth
\emph{below} chance on a randomized two-choice question rules out position bias and points to a
systematic inversion, not a prompt-wording artifact: rephrasing front/back as closer/farther from the
camera leaves accuracy below chance (Table~\ref{tab:depth_convention}).
\begin{wraptable}{r}{0.47\linewidth}
\vspace{-\baselineskip}
\centering
\caption{\textbf{Not a prompt-wording artifact.} Qwen2.5-VL-7B, depth $n{=}153$, chance $0.5$. Both
phrasings stay below chance, so unambiguous camera-relative wording does not recover depth.}
\label{tab:depth_convention}
\vspace{0.2em}\renewcommand{\arraystretch}{1.1}\setlength{\tabcolsep}{5pt}\small
\begin{tabular}{l c c}
\toprule
\textbf{Phrasing} & \textbf{acc.} & \textbf{vs.\ ch.} \\
\midrule
``front / back''        & 28.1 & $-21.9$ \\
``closer / farther''    & 32.7 & $-17.3$ \\
\bottomrule
\end{tabular}
\vspace{-0.5\baselineskip}
\end{wraptable}

\subhead{Decodability: every axis, vertical most.} A linear probe tells a very different story
(Fig.~\ref{fig:loci}). Decoding the within-axis pole from the decision-token residual
stream rises sharply at layers 16--24 for all three axes and peaks well above the shuffled-label
floor: horizontal $84.9$, depth $77.2$, and vertical $94.1$, the \emph{highest} of the three.
The image-pooled locus stays low everywhere ($59$--$67$), so the direction is assembled in the
residual stream at the decision/text locus in mid-to-late layers rather than read directly off image
tokens.

\subhead{The dissociation.} Placing behavior next to decodability (Table~\ref{tab:dissociation})
gives the gap that drives the rest of the paper. Horizontal is decoded \emph{and} deployed
(gap $+4$). Vertical is the most decodable axis yet behaves near chance (gap $+36$). Depth is
decodable while behavior runs \emph{below} chance (gap $+47$, anti-correlated). Taken on its own,
this is the canonical knows-but-does-not-deploy picture: a strong latent representation
(vertical $94\%$) that the model fails to act on, plus an inverted readout for depth. It is exactly
the configuration that invites a training-free recovery method, which we build next.

\begin{table}[t]
\centering
\caption{\textbf{Per-axis behavior vs.\ linear decodability} (Qwen2.5-VL-7B, ViewSpatial
Camera-Relative-Direction). \textbf{18-way} native accuracy (chance 25); \textbf{binary} within-axis
poles (chance 50); \textbf{decode} peak probe accuracy (shuffle floor in parentheses). The \textbf{gap}
(decode$-$binary) is small for horizontal, large for vertical, largest for depth where behavior is
\emph{below} chance. Accuracies in \%.}
\label{tab:dissociation}
\vspace{0.3em}
\renewcommand{\arraystretch}{1.15}
\setlength{\tabcolsep}{5pt}
\footnotesize
\begin{tabular}{l c c c c c l}
\toprule
\textbf{Axis} & \textbf{$n$} & \textbf{18-way} & \textbf{binary} & \textbf{decode (shuffle)} & \textbf{gap} & \textbf{behavioral verdict} \\
\midrule
Horizontal (L/R)   & 518 & 74.9 & \textbf{81.3} & 84.9 \,(57) & $+4$ & deployed, correct \\
Vertical (up/dn)   & 376 & 26.9 & 58.5 & \textbf{94.1} \,(63) & $+36$ & near chance \\
Depth (front/back) & 153 & 16.3 & 30.7 & 77.2 \,(55) & $+47$ & \emph{below} chance \\
\bottomrule
\end{tabular}
\end{table}

\subsection{An Intervention That Appears to Work}
\label{sec:intervention}

If vertical is decodable at $94\%$ but behaves at chance, the natural move is to inject the decoded
direction and watch behavior follow. It does.

\subhead{Causal steering.} Adding the per-axis diff-of-means direction at the peak layer cleanly and
monotonically moves the positive-pole probability for all three axes (Appendix~\ref{app:intervention},
Fig.~\ref{fig:causal_sweep}; depth moves the wrong way, confirming an inverted readout), establishing
causal control. A corrective oracle raises vertical from $59\%$ to $74\%$, and projecting the direction
\emph{out} collapses it to chance, a necessity signature. A multi-layer band around the decode peak
(L20--24) recovers far more than any single layer, locating the effect on a distributed path
(Appendix~\ref{app:robustness}, Table~\ref{tab:band}); the \emph{decoding}-optimal probe direction is
\emph{not} the intervention-optimal one (Table~\ref{tab:probe_vs_diffmean}).

\begin{figure}[t]
\centering
\begin{tikzpicture}
\begin{axis}[
    width=0.70\linewidth, height=3.2cm,
    xmin=-0.2, xmax=8.4, ymin=50, ymax=84,
    xtick={0,0.5,1,2,4,8},
    ytick={50,60,70,80},
    xlabel={\small projection strength $\gamma$},
    ylabel={\small vertical accuracy},
    tick label style={font=\footnotesize},
    ymajorgrids, grid style={gray!18},
    legend style={at={(0.5,1.04)}, anchor=south, legend columns=3,
                  font=\footnotesize, draw=neutralGray!50},
    legend cell align=left,
    mark size=2.0pt, thick,
]
\addplot[invertedVerm, mark=*] coordinates {(0,58.5)(0.5,63.3)(1,69.4)(2,79)(4,68.9)(8,58.8)};
\addlegendentry{in-band, real}
\addplot[priorSlate, mark=square*, densely dashed] coordinates {(0,59.6)(2,78.7)};
\addlegendentry{in-band, gray}
\addplot[accentBlue, mark=triangle*, dotted] coordinates {(0,58.5)(0.5,58.5)(1,58)(2,57.7)(4,55.9)(8,55.1)};
\addlegendentry{off-band, real}
\node[font=\tiny, neutralGray!90!black, anchor=west] at (axis cs:2.2,79) {gray measured at $\gamma{=}0,2$};
\end{axis}
\end{tikzpicture}
\caption{\textbf{The recovery is prior amplification, not visual deployment.} Training-free projection
along the localized vertical direction lifts vertical from $58.5$ to $79$ at $\gamma=2$ with horizontal
intact. But the \emph{same} projection with a gray-blank image (dashed) gains just as much ($+19.1$
vs.\ $+20.5$): the lever amplifies a non-visual prior. An off-band projection (L2--6, dotted) does
nothing. Percentages.}
\label{fig:projection}
\end{figure}
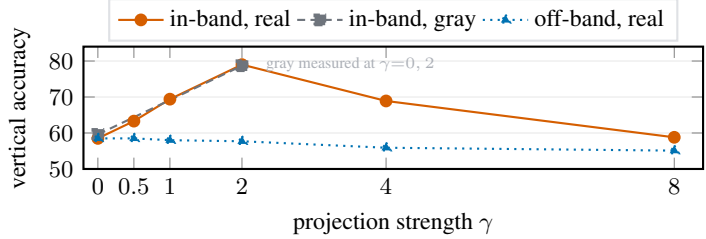

\subhead{Training-free projection.} The strongest version uses no gradients or labels: we amplify the
localized vertical signal with one global scalar $\gamma$ (\S\ref{sec:setup}). Vertical rises from
$59\%$ to $\mathbf{79\%}$ at $\gamma{=}2$ ($+21$) with horizontal essentially unchanged
($82\%\to81\%$), positive across a broad range, while an off-band projection does nothing
(Appendix~\ref{app:intervention}, Fig.~\ref{fig:projection}), so the lever is real and localized to the
decode band. Depth does \emph{not} lift cleanly, consistent with an inverted/entangled mode rather than
an under-deployed one.

\noindent\textit{The result we could have stopped at.} Vertical is the most decodable axis ($94\%$),
behaves at chance ($59\%$), and is recovered to $79\%$ by a training-free, label-free projection that
leaves horizontal intact, while depth fails to recover, a clean dissociation. Reported as is, this
reads as latent vertical spatial knowledge recovered without training. \emph{It would be wrong.}

\subsection{The Vision-Ablation Arbiter}
\label{sec:arbiter}

The recovery in \S\ref{sec:intervention} never checked whether the behavior it moved depends on the
image. We run that check by re-evaluating every condition with the image replaced by a same-size gray
blank (\S\ref{sec:setup}). Numbers here are for Qwen2.5-VL-7B; \S\ref{sec:crossmodel} shows the regimes
recur across fourteen models.

\subhead{The arbiter is discriminating, not deflationary.} The three axes respond to blinding in
three predictable ways (Table~\ref{tab:arbiter}). \textbf{Horizontal} collapses from $82\%$ to chance
($51\%$ balanced): its correct answers were image-driven, so the arbiter is a valid positive control.
\textbf{Depth} sits \emph{below} chance with the image ($31\%$) but \emph{rises} to chance when blinded
($45\%$): the inversion itself is caused by the image; remove the input and the systematic error
disappears. \textbf{Vertical} barely moves ($59\%\to60\%$ raw, $70\%\approx70\%$ balanced): the model
is not using the image for vertical at all. Four further ablations (black, noise, patch-scramble, and a
\emph{real but mismatched} in-distribution image) give the same verdict
(Appendix~\ref{app:robustness}, Table~\ref{tab:robustness}), so the taxonomy is a property of how the
model uses the image, not an artifact of the blank.

\begin{table}[t]
\centering
\caption{\textbf{The vision-ablation arbiter and the three grounding regimes} (Qwen2.5-VL-7B). Per axis:
accuracy with the \textbf{real} image and a \textbf{gray} blank (raw and balanced), and the
training-free projection \textbf{gain} with the real vs.\ gray image. The regime is read off the
\emph{real$-$gray} baseline gap and \emph{whether the gain needs the image}. Colors:
\colorbox{rowcatE}{grounded+correct}, \colorbox{rowcatD}{grounded+inverted},
\colorbox{rowcatB}{prior}. Accuracies in \%.}
\label{tab:arbiter}
\vspace{0.3em}
\renewcommand{\arraystretch}{1.2}
\setlength{\tabcolsep}{5pt}
\small
\begin{tabular}{l cc cc cc l}
\toprule
& \multicolumn{2}{c}{\textbf{baseline (raw)}} & \multicolumn{2}{c}{\textbf{baseline (balanced)}}
& \multicolumn{2}{c}{\textbf{proj.\ gain}} & \\
\cmidrule(lr){2-3}\cmidrule(lr){4-5}\cmidrule(lr){6-7}
\textbf{Axis} & real & gray & real & gray & real & gray & \textbf{grounding regime} \\
\midrule
\rowcolor{rowcatE}
Horizontal & 82.4 & \textbf{49.4} & 82.6 & \textbf{50.7} & $+3$ & $\sim$0 & vision-dep.\ \& correct \\
\rowcolor{rowcatD}
Depth      & \textbf{31.4} & 45.1 & \textbf{31.6} & 46.3 & $\le$0 & $\sim$0 & vision-dep.\ \& inverted \\
\rowcolor{rowcatB}
Vertical   & 58.5 & 59.6 & 69.8 & 70.3 & $+21$ & $+19$ & vision-\emph{indep.} (prior) \\
\bottomrule
\end{tabular}
\vspace{0.3em}\\
\footnotesize
Reading. \textbf{Horizontal}: removing the image collapses accuracy to chance ($82\!\to\!49$),
and the projection gain vanishes with the gray image, so the signal is genuinely visual. \textbf{Depth}:
real accuracy is \emph{below} chance but the gray image reverts to chance ($31\!<\!45$): the
inversion is caused by looking at the image. \textbf{Vertical}: real $\approx$ gray under both raw and
balanced scoring, and the projection gain is the \emph{same} with or without the image, so the model is
not using vision and the recovery amplifies a linguistic prior. Back-solving the per-class recalls on
the $31/69$ pole split gives recall$(\text{above})\!\approx\!99$, recall$(\text{below})\!\approx\!40$:
a directional default-to-above response bias, not noise.
\end{table}

\subhead{The recovery amplifies a prior.} The decisive test is whether the projection gain of
\S\ref{sec:intervention} needs the image. It does not: with the gray blank the vertical projection
gains $+19$ at $\gamma{=}2$, indistinguishable from the $+21$ it gains with the real image (balanced:
$+12$ blind vs.\ $+14$ sighted). The lever is a direction that pushes the output distribution, and that
distribution was never conditioned on the image. The horizontal control behaves oppositely: its small
already-deployed gain vanishes under the blank, as there is no prior to amplify. What vertical encodes
is a directional default: balanced accuracy ($70\%$) exceeds raw ($59\%$) on the $31/69$ above/below
split, with recall$(\text{above})\approx99\%$ and recall$(\text{below})\approx40\%$. The probe decodes
vertical at $94\%$ because this default-to-above is linearly readable, a language-side prior, the
residual-stream analogue of a blind-VQA prior \citep{goyal2017vqa, agrawal2018dontassume} that probe
accuracy cannot distinguish from grounded knowledge \citep{hewitt2019control, belinkov2022probing}.

\subhead{Consequence.} The headline of \S\ref{sec:intervention} does not survive: vertical is a
prior, not under-deployed visual knowledge, and depth is inverted rather than absent. Depth also
resists a clean training-free flip: sweeping a signed projection at both signs, linear edits only
neutralize the inversion toward chance (best $\approx47\%$ vs.\ the $\approx68\%$ a global sign-flip
would give), revealing an \emph{entangled} mode rather than a single free sign; a minimal \emph{trained}
edit at the causally localized layer does re-deploy it (\S\ref{sec:method}). On this model, probing and
steering overstate grounded spatial knowledge, and one cheap control says by how much and where.

\begin{table}[t]
\centering
\caption{\textbf{The arbiter is robust to the input-degeneracy objection} (Qwen2.5-VL-7B, balanced
accuracy, full $n$, Run 12). Five vision ablations, including \texttt{mismatch}: a real-but-wrong
\emph{same-axis} image (fully in-distribution). All five agree with the gray-blank verdict. All values are balanced accuracy in \%.}
\label{tab:robustness}
\vspace{0.3em}\renewcommand{\arraystretch}{1.15}\setlength{\tabcolsep}{6pt}\small
\begin{tabular}{l c | c c c c c | l}
\toprule
\textbf{axis} & \textbf{real} & \textbf{gray} & \textbf{black} & \textbf{noise} & \textbf{scramble} & \textbf{mismatch} & \textbf{verdict} \\
\midrule
Horizontal & 82.6 & 50.7 & 49.6 & 47.4 & 54.1 & 50.7 & all $\to$ chance (grounded, correct) \\
Vertical   & 69.8 & 70.3 & 70.1 & 66.5 & 60.5 & 68.6 & all $\approx$ real (prior) \\
Depth      & 31.6 & 46.3 & 44.2 & 42.7 & 48.9 & 45.7 & all $\to$ chance from below (inverted) \\
\bottomrule
\end{tabular}
\vspace{0.2em}\\
\footnotesize The \texttt{mismatch} column is the decisive in-distribution control: a correct,
real, same-axis image that simply does not match the question. Horizontal still collapses ($50.7$),
vertical is unchanged ($68.6\approx69.8$, so it ignores image \emph{content}, not just image
\emph{presence}), and depth's inversion relaxes to chance ($45.7$). The taxonomy is not an artifact
of feeding the model a degenerate blank.
\end{table}

\section{The Taxonomy Is Architecture-General and Task-Specific}
\label{sec:crossmodel}

The single-model arc could be a Qwen2.5-VL-7B idiosyncrasy. We run the independent battery (per-axis
behavior, the vision arbiter, and a per-layer probe) on thirteen further VLMs, for fourteen models
across six LM families (Qwen2/2.5/3, Mistral, SmolLM, Gemma) and 2B--27B, including three within-family
scale ladders (full roster in Appendix~\ref{app:crossmodel}, Table~\ref{tab:xmodel}; matrix in
Fig.~\ref{fig:cardgrid}).

\subhead{The three regimes recur across LM families.} In every capable model horizontal is
vision-dependent and correct (real $\gg$ blank $\to$ chance), vertical is prior-dominated (real
$\approx$ blank), and depth is at or below chance. This holds for \textbf{Pixtral-12B} (Mistral LM) and
\textbf{Gemma-3} (a different family with a SigLIP rather than Qwen-derived vision tower), so the
taxonomy is a property of current VLMs, not a Qwen or InternViT artifact. \textbf{SmolVLM-2.2B} is the one all-axis
exception, near chance even on horizontal ($47\%$): a capability floor below which no axis is grounded,
exactly as expected.

\subhead{Decodability overstates knowledge in every model.} The probe decodes vertical at
$87$--$97\%$ and depth at $73$--$86\%$ in every capable model while behavior on those axes is prior or
inverted. This gap between what the probe reads and what the model grounds is systematic across
architectures, and the cautionary claim of \S\ref{sec:arbiter} is the paper's most robust result.

\subhead{Depth inversion is scale-emergent within families.} Three within-family ladders, varying only
LM scale, show the inversion \emph{emerging} with size (Appendix~\ref{app:crossmodel},
Fig.~\ref{fig:depth_scale}): InternVL3 ($38\%\to31\%\to28\%$) and Qwen2.5-VL ($48\%\to32\%$) deepen
monotonically, and in Gemma-3 ($53\%\to43\%\to46\%$) it emerges by 12B with horizontal grounding also
emerging ($50\%\to67\%$). Onset and severity are family-modulated (InternVL3 is inverted already at
2B, Qwen2.5-VL-3B and Gemma-4B not yet, and Gemma-3 partially recovers at 27B), so the law is that
inversion emerges with LM scale within a family, with degree and onset family-specific.

\subhead{The LM backbone sets the fingerprint.} LLaVA-OneVision-7B and InternVL3-8B share the Qwen2-7B
LM with different vision towers (SigLIP vs.\ InternViT) and are qualitatively identical on all three
axes, differing only in the magnitude of the weak vertical vision signal. The grounding fingerprint is
thus driven by the language backbone with the vision tower and scale tuning the degree, consistent
with the locus analysis (Appendix~\ref{app:layers}, Fig.~\ref{fig:loci}), where the axis signal is
assembled LM-side.

\begin{figure}[t]
\centering
\begin{tikzpicture}[
    cell/.style={rectangle, minimum width=8.5mm, minimum height=8.5mm,
                 inner sep=0pt, font=\tiny, draw=neutralGray!60, line width=0.3pt},
    collbl/.style={font=\scriptsize, anchor=south west, rotate=45},
    rowlbl/.style={font=\scriptsize, anchor=east},
    legcap/.style={font=\scriptsize, anchor=west, neutralGray!90!black},
]
\def\colstep{0.88}

\newcommand{\cardcol}[8]{%
  \node[collbl] at (\colstep*#1 - 0.15, 0.5) {#2};
  \node[cell, fill=#3!28!white, draw=#3!80!black] at (\colstep*#1,  0) {\textbf{#4}};
  \node[cell, fill=#5!28!white, draw=#5!80!black] at (\colstep*#1, -0.88) {\textbf{#6}};
  \node[cell, fill=#7!28!white, draw=#7!80!black] at (\colstep*#1, -1.76) {\textbf{#8}};
}

\node[rowlbl] at (-0.05, 0)     {horizontal};
\node[rowlbl] at (-0.05, -0.88) {vertical};
\node[rowlbl] at (-0.05, -1.76) {depth};

\cardcol{1}{Qwen2.5-VL-7B}{groundedTeal}{83}{priorSlate}{59}{invertedVerm}{31}
\cardcol{2}{Qwen3-VL-8B}{groundedTeal}{88}{priorSlate}{95}{invertedVerm}{22}
\cardcol{3}{Pixtral-12B}{groundedTeal}{71}{priorSlate}{67}{invertedVerm}{38}
\cardcol{4}{Qwen2.5-VL-3B}{groundedTeal}{67}{priorSlate}{66}{neutralGray}{48}
\cardcol{5}{InternVL3-2B}{neutralGray}{47}{neutralGray}{57}{invertedVerm}{38}
\cardcol{6}{InternVL3-8B}{groundedTeal}{88}{priorSlate}{62}{invertedVerm}{31}
\cardcol{7}{InternVL3-14B}{groundedTeal}{85}{priorSlate}{64}{invertedVerm}{28}
\cardcol{8}{Gemma-3-4B}{groundedTeal}{60}{priorSlate}{60}{neutralGray}{53}
\cardcol{9}{Gemma-3-12B}{groundedTeal}{80}{priorSlate}{61}{invertedVerm}{43}
\cardcol{10}{Gemma-3-27B}{groundedTeal}{82}{priorSlate}{62}{invertedVerm}{46}
\cardcol{11}{LLaVA-NeXT-7B}{groundedTeal}{73}{priorSlate}{61}{invertedVerm}{42}
\cardcol{12}{SpaceQwen-3B}{groundedTeal}{69}{priorSlate}{61}{invertedVerm}{45}
\cardcol{13}{SpaceOm-3B}{groundedTeal}{67}{priorSlate}{60}{invertedVerm}{46}
\cardcol{14}{SmolVLM-2.2B}{neutralGray}{47}{neutralGray}{57}{neutralGray}{48}

\def\lgy{-2.95}
\node[cell, fill=groundedTeal!28!white, draw=groundedTeal!80!black,
      minimum width=6mm, minimum height=4mm] at (0.7,\lgy) {};
\node[legcap] at (1.05,\lgy) {grounded};

\node[cell, fill=priorSlate!28!white, draw=priorSlate!80!black,
      minimum width=6mm, minimum height=4mm] at (3.5,\lgy) {};
\node[legcap] at (3.85,\lgy) {prior};

\node[cell, fill=invertedVerm!28!white, draw=invertedVerm!80!black,
      minimum width=6mm, minimum height=4mm] at (5.5,\lgy) {};
\node[legcap] at (5.85,\lgy) {inverted};

\node[cell, fill=neutralGray!28!white, draw=neutralGray!80!black,
      minimum width=6mm, minimum height=4mm] at (8.0,\lgy) {};
\node[legcap] at (8.35,\lgy) {chance / floor};

\end{tikzpicture}
\caption{\textbf{The grounding taxonomy as a cross-model card.} Real-image accuracy on the three
ViewSpatial axes for 14 VLMs (six LM families, 2B--27B), each cell colored by its arbiter regime. At a
glance: horizontal grounded, vertical a prior, depth inverted in every capable model, with depth
inversion scale-emergent within families (InternVL3 $38\!\to\!31\!\to\!28$; Gemma non-monotone
$53\!\to\!43\!\to\!46$); SmolVLM-2.2B is a capability floor. Percentages.}
\label{fig:cardgrid}
\end{figure}

\subsection{External Benchmarks: A Task-Type Boundary}
\label{sec:external}

A single-benchmark taxonomy invites the objection that ViewSpatial is simply broken on the
non-horizontal axes. We test this by running the behavioral battery and arbiter on two external
benchmarks (What'sUp \citep{kamath2023whatsup}, tabletop on/under and front/behind; and 3DSRBench
\citep{ma20243dsrbench}, real-scene height and closer-to-camera), mapping their categories onto our axes
Table~\ref{tab:external}. The result replicates the taxonomy and
sharpens it into a task-type boundary.

\subhead{Depth inversion is camera-egocentric, not a general depth failure.} The decisive result is a
dissociation \emph{within} the same model (Appendix~\ref{app:crossmodel}, Table~\ref{tab:external}): Qwen2.5-VL-7B inverts depth on
ViewSpatial's camera-relative front/back yet is grounded-correct on What'sUp near-field front/behind
and 3DSRBench closer-to-camera, with mismatch controls confirming both are vision-driven; four further
models show the same pattern, and horizontal replicates cleanly everywhere as a positive control. The
inversion is thus a reproducible property of
\emph{camera-egocentric front/back} framing, a task type rather than a missing capability or a benchmark
artifact, which pre-empts the artifact objection by exhibiting the contrast in one model. Vertical is
likewise task-type-dependent (grounded on tabletop on/under, a prior on ViewSpatial above/below and
3DSRBench height): grounding is organized by task type, with simple near-field contact relations
grounded while abstract camera-perspective and real-scene relations fall back to priors or invert, and
the arbiter transfers unchanged to make this legible.

\section{Correcting the Inversion Diagnoses Its Geometry}
\label{sec:method}

If the depth axis is decodable and causally controllable (\S\ref{sec:intervention}) but read out with
the wrong sign, what is the \emph{minimal} edit that re-deploys it? We run a battery of nine
corrections (scalar recalibration, logit-lens reflection, a norm-preserving angular rotation
\citep{angular2025}, selective multi-layer rotation, activation edits SEA \citep{sea2024} and SADI
\citep{sadi2024}, per-head reflection ITI \citep{iti2023}, SAE feature steering
\citep{pach2025saevlm}, and a trained low-rank edit LoRA \citep{lora2021} / ReFT \citep{wu2024reft})
on four models, then test generalization on four more (Appendix~\ref{app:correction}). Two findings
emerge: the decode$\neq$deploy inversion is a \emph{population} phenomenon, and the minimal correction
is a per-model \emph{complexity spectrum}.

\subhead{The inversion is a population phenomenon.} A signed-slope diagnostic (the sign of the steering
response of decodability versus behavior along depth) finds the inverted configuration (decode-slope
$>0$, deploy-slope $<0$) on \textbf{seven of eight inverted-capable models across five LM families}
(Fig.~\ref{fig:forest}, Table~\ref{tab:population}); the only exception is the not-yet-inverted scale
floor Qwen2.5-VL-3B. Decodability rises with the steer everywhere while behavior moves the opposite way,
ruling out a single-model artifact.

\subhead{The minimal correction is a per-model complexity spectrum.} \emph{Which} correction
re-deploys the axis reads out each model's geometry. On Qwen3-VL-8B, the cleanest case, a
training-free, norm-preserving rotation by $\theta\approx\pi$ nearly triples depth accuracy
($21\%\to77\%$) and, across three seeds, \emph{matches} a trained low-rank edit ($+49{\pm}8$ vs.\
$+55{\pm}2$): the predicted un-inversion with no training. Across the population, \emph{which rung suffices} is itself the
diagnostic: a clean rotation on Qwen3-VL-8B, a partial one on LLaVA-OV ($+14$) and InternVL3-8B ($+8$),
recalibration on InternVL3-14B, a trained edit on the distributed Qwen2.5-VL-7B ($+43$), and
suppress-only (decodable but not installable) on InternVL3-2B and Pixtral-12B. The minimal
sufficient correction ladders with the geometry (recalibration $\to$ rotation $\to$ trained low-rank
$\to$ none), and the winning rung is a per-model fingerprint (Fig.~\ref{fig:method}; full battery in
Appendix~\ref{app:correction}, Table~\ref{tab:method}).

\subhead{The causal locus is mid-late, one-sided, and shared across the population.} Localizing the
inverted channel layer-by-layer in seven models (Fig.~\ref{fig:locality2}), the causal-controllability
peak sits mid-late (L19--36), at or downstream of the decode peak and \emph{never} early, and is
strongly \emph{asymmetric} (one steer sign recovers correct depth, the opposite does almost
nothing) in 6/7 models; its magnitude tracks correction-complexity ($0.99$ for Qwen3-VL-8B down to
$\le0.56$ for trained-only models), so localization and correction cross-validate. On Qwen3-VL-8B
(Appendix~\ref{app:correction}, Table~\ref{tab:model2}) the correctable channel is low-dimensional yet
off the probe axis (a rank-1 ReFT $+58$ and a 4-component ICA reflection $+60$ recover it; the single
probe direction and a GEVD subspace fail, $+0$); the training-free$=$trained tie is seed-stable. This closes the arc from diagnosis through mechanism to a geometry-diagnostic
correction: a shared inversion whose minimal fix, per model, reads out its geometry.

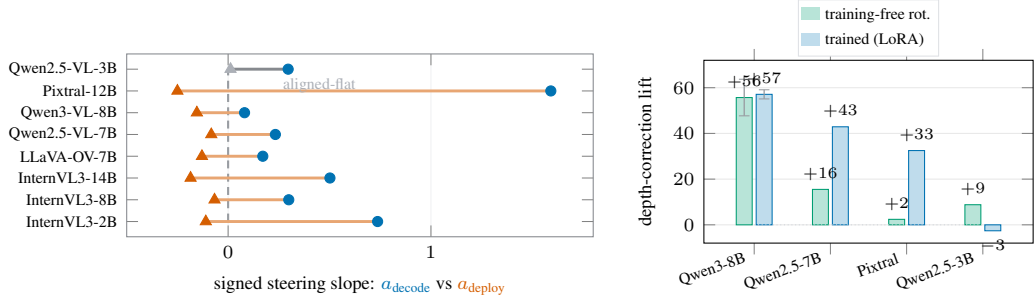
\begin{figure}[t]
\centering
\begin{subfigure}[t]{0.555\linewidth}
\centering
\begin{tikzpicture}
\begin{axis}[
    width=\linewidth, height=4.0cm,
    xmin=-0.5, xmax=1.8,
    xlabel={\scriptsize signed steering slope: \textcolor{accentBlue}{$a_{\text{decode}}$} vs \textcolor{invertedVerm}{$a_{\text{deploy}}$}},
    ytick=data,
    symbolic y coords={iv2b,iv8b,iv14b,llava,qwen,qwen3,pixtral,q3b},
    yticklabels={InternVL3-2B, InternVL3-8B, InternVL3-14B, LLaVA-OV-7B, Qwen2.5-VL-7B, Qwen3-VL-8B, Pixtral-12B, Qwen2.5-VL-3B},
    yticklabel style={font=\tiny},
    tick label style={font=\scriptsize},
    xtick={0,1},
    xmajorgrids, grid style={neutralGray!25, very thin},
    axis line style={neutralGray!70!black},
    clip=false, enlarge y limits=0.10,
]
\draw[neutralGray!80!black, densely dashed, thick] (axis cs:0,iv2b) -- (axis cs:0,q3b);
\draw[invertedVerm!55, line width=1.1pt] (axis cs:0.737,iv2b) -- (axis cs:-0.110,iv2b);
\draw[invertedVerm!55, line width=1.1pt] (axis cs:0.299,iv8b) -- (axis cs:-0.067,iv8b);
\draw[invertedVerm!55, line width=1.1pt] (axis cs:0.502,iv14b) -- (axis cs:-0.185,iv14b);
\draw[invertedVerm!55, line width=1.1pt] (axis cs:0.171,llava) -- (axis cs:-0.129,llava);
\draw[invertedVerm!55, line width=1.1pt] (axis cs:0.233,qwen) -- (axis cs:-0.083,qwen);
\draw[invertedVerm!55, line width=1.1pt] (axis cs:0.081,qwen3) -- (axis cs:-0.154,qwen3);
\draw[invertedVerm!55, line width=1.1pt] (axis cs:1.591,pixtral) -- (axis cs:-0.250,pixtral);
\draw[neutralGray!70!black, line width=1.1pt] (axis cs:0.296,q3b) -- (axis cs:0.012,q3b);
\addplot[only marks, mark=*, mark size=1.9pt, accentBlue] coordinates {
    (0.737,iv2b)(0.299,iv8b)(0.502,iv14b)(0.171,llava)(0.233,qwen)(0.081,qwen3)(1.591,pixtral)(0.296,q3b)};
\addplot[only marks, mark=triangle*, mark size=2.4pt, invertedVerm] coordinates {
    (-0.110,iv2b)(-0.067,iv8b)(-0.185,iv14b)(-0.129,llava)(-0.083,qwen)(-0.154,qwen3)(-0.250,pixtral)};
\addplot[only marks, mark=triangle*, mark size=2.4pt, neutralGray!90!black] coordinates {(0.012,q3b)};
\node[anchor=north, font=\tiny, neutralGray!90!black] at (axis cs:0.45,q3b) {aligned-flat};
\end{axis}
\end{tikzpicture}
\caption{Population: $a_{\text{decode}}{>}0$ but $a_{\text{deploy}}{<}0$ (inverted) in 7/8 models.}
\label{fig:forest}
\end{subfigure}\hfill
\begin{subfigure}[t]{0.425\linewidth}
\centering
\begin{tikzpicture}
\begin{axis}[
    width=\linewidth, height=4.0cm,
    ybar=1.5pt, bar width=6pt,
    enlarge x limits=0.22,
    ymin=-8, ymax=72,
    ytick={0,20,40,60},
    ymajorgrids, grid style={gray!18},
    ylabel={\scriptsize depth-correction lift},
    symbolic x coords={Qwen3-8B, Qwen2.5-7B, Pixtral, Qwen2.5-3B},
    xtick=data, xticklabel style={font=\tiny, rotate=20, anchor=east},
    tick label style={font=\scriptsize},
    legend style={at={(0.5,1.02)}, anchor=south, legend columns=1,
                  font=\tiny, draw=neutralGray!50, inner sep=1pt},
    legend cell align=left,
    legend image code/.code={\draw[#1, draw=none] (0cm,-0.06cm) rectangle (0.25cm,0.10cm);},
    nodes near coords, nodes near coords style={font=\tiny, /pgf/number format/.cd, fixed, precision=0, print sign},
]
\addplot[draw=groundedTeal, fill=groundedTeal!28!white,
         error bars/.cd, y dir=both, y explicit, error bar style={neutralGray!80!black, line width=0.5pt}]
    coordinates {(Qwen3-8B,55.7) +-(0,8) (Qwen2.5-7B,15.5) (Pixtral,2.4) (Qwen2.5-3B,8.8)};
\addlegendentry{training-free rot.}
\addplot[draw=accentBlue, fill=accentBlue!25!white,
         error bars/.cd, y dir=both, y explicit, error bar style={neutralGray!80!black, line width=0.5pt}]
    coordinates {(Qwen3-8B,57.1) +-(0,2) (Qwen2.5-7B,42.9) (Pixtral,32.5) (Qwen2.5-3B,-2.6)};
\addlegendentry{trained (LoRA)}
\draw[neutralGray!70, densely dotted] (axis cs:Qwen3-8B,0) -- (axis cs:Qwen2.5-3B,0);
\end{axis}
\end{tikzpicture}
\caption{Correction spectrum: training-free rotation matches trained LoRA only on Qwen3-VL-8B.}
\label{fig:method}
\end{subfigure}
\caption{\textbf{The decode$\neq$deploy inversion is a population phenomenon (left); which minimal edit
re-deploys it reads out each model's geometry (right).} \emph{Left:} signed steering slope of
\emph{decodability} ($a_{\text{decode}}$, blue) and \emph{behaviour} ($a_{\text{deploy}}$, vermillion)
along depth; decodability rises everywhere while behaviour moves the opposite way in 7/8 models across
five LM families (exception: the not-yet-inverted Qwen2.5-VL-3B). \emph{Right:} depth-accuracy lift over
baseline. On the clean-rotation Qwen3-VL-8B a training-free rotation matches a trained LoRA across three seeds
($+49{\pm}8$ vs.\ $+55{\pm}2$); on distributed (Qwen2.5-VL-7B) and not-installable (Pixtral-12B)
inversions it does not, so the winning rung is a per-model fingerprint. Percentages.}
\label{fig:popcorr}
\end{figure}
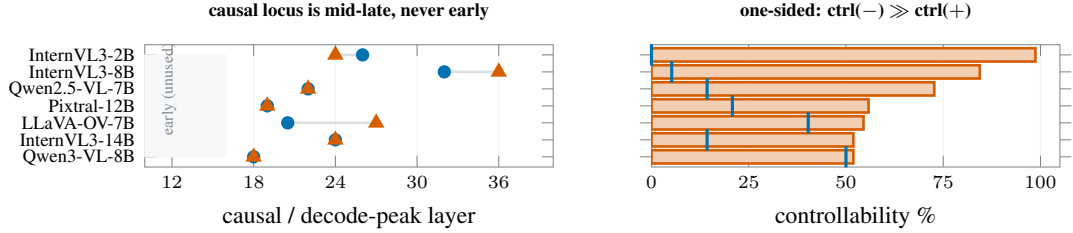
\begin{figure}[t]
\centering
\begin{tikzpicture}
\begin{groupplot}[
    group style={group size=2 by 1, horizontal sep=1.3cm},
    height=3.2cm,
    symbolic y coords={iv2b,iv8b,qwen,pixtral,llava,iv14b,qwen3},
    ytick=data,
    tick label style={font=\scriptsize},
    every axis plot/.append style={line width=1.0pt},
    axis line style={neutralGray!70!black},
]
\nextgroupplot[
    width=0.50\linewidth,
    xmin=10, xmax=40, xtick={12,18,24,30,36},
    xlabel={\small causal / decode-peak layer},
    yticklabels={InternVL3-2B, InternVL3-8B, Qwen2.5-VL-7B, Pixtral-12B, LLaVA-OV-7B, InternVL3-14B, Qwen3-VL-8B},
    xmajorgrids, grid style={neutralGray!20, very thin},
    title={\scriptsize\bfseries causal locus is mid-late, never early},
    enlarge y limits=0.10,
]
\draw[neutralGray!12, fill=neutralGray!12] (axis cs:10,iv2b) rectangle (axis cs:16,qwen3);
\node[font=\tiny, neutralGray!70!black, anchor=south, rotate=90] at (axis cs:13,llava) {early (unused)};
\draw[neutralGray!55, line width=1.0pt] (axis cs:26,qwen3)--(axis cs:24,qwen3);
\draw[neutralGray!55, line width=1.0pt] (axis cs:32,iv14b)--(axis cs:36,iv14b);
\draw[neutralGray!55, line width=1.0pt] (axis cs:22,llava)--(axis cs:22,llava);
\draw[neutralGray!55, line width=1.0pt] (axis cs:19,pixtral)--(axis cs:19,pixtral);
\draw[neutralGray!55, line width=1.0pt] (axis cs:20.5,qwen)--(axis cs:27,qwen);
\draw[neutralGray!55, line width=1.0pt] (axis cs:24,iv8b)--(axis cs:24,iv8b);
\draw[neutralGray!55, line width=1.0pt] (axis cs:18,iv2b)--(axis cs:18,iv2b);
\addplot[only marks, mark=*, mark size=2.0pt, accentBlue] coordinates {
    (26,qwen3)(32,iv14b)(22,llava)(19,pixtral)(20.5,qwen)(24,iv8b)(18,iv2b)};
\addplot[only marks, mark=triangle*, mark size=2.6pt, invertedVerm] coordinates {
    (24,qwen3)(36,iv14b)(22,llava)(19,pixtral)(27,qwen)(24,iv8b)(18,iv2b)};
\legend{}

\nextgroupplot[
    width=0.50\linewidth,
    xmin=0, xmax=105, xtick={0,25,50,75,100},
    xlabel={\small controllability \%},
    yticklabels={,,,,,,},
    xmajorgrids, grid style={neutralGray!20, very thin},
    title={\scriptsize\bfseries one-sided: ctrl($-$) $\gg$ ctrl($+$)},
    enlarge y limits=0.10,
    xbar, bar width=5pt,
]
\addplot[xbar, draw=invertedVerm, fill=invertedVerm!30!white] coordinates {
    (98.7,qwen3)(84.4,iv14b)(72.7,llava)(55.8,pixtral)(54.5,qwen)(51.9,iv8b)(51.9,iv2b)};
\addplot[only marks, mark=|, mark size=4pt, accentBlue, line width=1.2pt] coordinates {
    (0.0,qwen3)(5.2,iv14b)(14.3,llava)(20.8,pixtral)(40.3,qwen)(14.3,iv8b)(50.0,iv2b)};
\legend{}
\end{groupplot}
\end{tikzpicture}
\caption{\textbf{The inverted depth channel is causal and localized across the population, with a
one-sided directional signature.} Seven inverted models. \emph{Left}: the probe-decode-peak (blue) and
causal-controllability-peak (vermillion) layers both sit mid-late, never early (shaded). \emph{Right}:
causal controllability is strongly asymmetric, the ctrl($-$) steer (bar) recovers correct depth while
ctrl($+$) (tick) does almost nothing, in 6/7 models; magnitude tracks card-C fixability. Controllability
in \%; left axis is layer index.}
\label{fig:locality2}
\end{figure}

\section{Discussion and Conclusion}
\label{sec:discussion}

\subhead{Decodability is not grounding, and steering recovery is not either.} Both halves of the
standard latent-knowledge pipeline failed to flag the vertical prior: the probe called it the \emph{most}
present axis ($94\%$) and the projection the \emph{most} recoverable, yet neither measures whether
behavior is conditioned on the image, so their agreement is not corroboration. This sharpens the
actionability-gap caution of \citet{basu2026interp}: a behavioral gain is not grounded recovery without
a control.

\subhead{The arbiter as a default control, and why three regimes.} The vision ablation is cheap,
assumption-light, and self-calibrating: it drives the genuinely visual horizontal axis to chance, so a
null on another axis is informative rather than an input artifact. A two-way grounded-vs-not split
would lump depth with the failures; the arbiter separates depth (vision-dependent, wrong sign) from
vertical (vision-independent). We recommend reporting real-vs-ablated accuracy and treating a gain
that survives ablation as prior amplification. Probing and steering overstate grounded knowledge because
they are blind to image-dependence; the arbiter is not.

\subhead{Why the inversion arises.} Four convergent findings point to a learned frame-convention
conflict rather than random damage. The inversion is \emph{construct-bound} (grounded-correct on
external depth, inverted only on ViewSpatial camera-relative front/back; Table~\ref{tab:external}),
\emph{scale-emergent} (deepens with LM scale, so learned), \emph{image-driven} (not a fixed text prior;
Appendix~\ref{app:robustness}), and a \emph{geometric mis-rotation} (a norm-preserving rotation
re-deploys it). The convergence indicates a depth readout learned against ViewSpatial's camera-relative
convention; tracing it to training data is future work.

\subhead{Limitations.} Depth-inversion emerges with LM scale but not monotonically; SmolVLM-2.2B bounds
the horizontal-universal claim to capable models; and our causal evidence is representation-level, as a
clean pixel-level minimal pair for camera-relative front/back is unattainable (a mirror is identity for
front/back, with no same-image pairs).

\section*{Reproducibility Statement}
All experiments use publicly available open-weight VLMs and the public ViewSpatial-Bench, What'sUp,
and 3DSRBench datasets. The task/axis decomposition, behavioral protocols (18-way and binary), probe
(five-fold CV logistic on PCA-50 decision-token features with a shuffled-label floor), the
training-free projection, the vision-ablation arbiter and its five image controls, the cross-model
battery, and the nine-method correction battery are specified in \S\ref{sec:setup} and
Appendix~\ref{app:additional}; per-model layers, hyperparameters, seeds, and the multi-seed protocol
for the rotation/trained-edit comparison are reported with the corresponding results
(\S\ref{sec:method}, Appendix~\ref{app:correction}). Limitations of single-seed and single-anchor
analyses are stated explicitly in \S\ref{sec:discussion}.

\bibliographystyle{plainnat}
\bibliography{bib/references}

\appendix
\section{Implementation Details}
\label{app:impl}

\paragraph{Model and decoding.} Qwen2.5-VL-7B-Instruct \citep{qwen2vl}, loaded in 4-bit nf4 on a
single RTX 4090, \texttt{attn\_implementation="sdpa"} (flash-attention not installed), greedy
decoding, \texttt{max\_new\_tokens}=8. The language tower has 28 decoder layers, hidden size 3584;
the image token id is 151655. We index \texttt{hidden\_states[$i$]} as the output of decoder layer
$i{-}1$. For Qwen2.5-VL under \texttt{transformers} 4.57 the decoder path is
\texttt{model.model.language\_model.layers}; we use a multi-path fallback to remain robust to
version changes.

\paragraph{Answer reading.} Letters are matched robustly over the variants \texttt{"A"}/\texttt{" A"}
to avoid tokenization-induced baseline drift. In the binary protocol the positive-pole letter is
randomized per item; all scoring, steering targets, and projections use the per-item
positive-pole map, and we verify the evaluation reproduces canonical baselines
(V $0.585$ / H $0.824$ / D $0.314$) before trusting any intervention number.

\paragraph{Probing.} Decision-token residual-stream activations are cached for all layers
($L{=}29$ including the embedding, $d{=}3584$) over the $1047$ pure-axis items. Per axis and layer we
fit a five-fold cross-validated logistic regression on PCA-50 features; the shuffled-label probe
gives the chance floor reported in brackets. We also report a light $|d'|$ sparse-neuron localizer on
decision-token activations; cross-axis top-20 neuron overlap is near zero, indicating distinct codes.

\paragraph{Steering scale.} Raw unit-vector steering ($\alpha\hat{d}$, $\alpha\le8$) is negligible
against residual-stream norms of $170$--$245$ and produces null effects. We therefore scale by the
standard deviation of the projection of activations onto $\hat{d}_l$, so $\alpha$ counts standard
deviations; we cap $|\alpha|\le4$ because larger values over-steer (at $|\alpha|{=}6$ the injected
norm reaches $\sim$65\% of $\lVert h\rVert$). Causal-control claims rest on the monotonicity of
$P(+\text{pole})$, not on accuracy at extreme $\alpha$.

\paragraph{Training-free projection.} On the $k{=}64$ highest-$|d'|$ neurons in the localized band we
add $\gamma\,(\sum_i \mathrm{sign}(d'_i)\,a_i(x))\,\hat{d}_l$ to the down-projection output, with a
single global $\gamma$ and no gradient updates. The band is L20--24 (vertical L21--25, depth L19--23);
the off-band control uses L2--6. We note that $\gamma$ is currently swept on the evaluation set; a
held-out $\gamma$ selection is the correct protocol and we report the broad positive range rather
than a single value to mitigate this.

\paragraph{Arbiter.} The gray-blank image is constructed at the same resolution as the real image and
passed through the identical preprocessing path (\texttt{make\_prompt\_blank}), so the only change is
pixel content. Baseline and projection are re-evaluated under this input with all else fixed.

\section{Additional Results}
\begin{table}[t]
\centering
\caption{\textbf{Capability matrix against interpretability studies of VLM spatial failure.} Columns:
representation treated as \emph{present}; grounding tested \emph{behaviorally} (vision ablation) vs.\
another probe; a \emph{sign-inversion} identified; a \emph{3-way} grounded/prior/inverted taxonomy; a
cross-model \emph{correction} spectrum; \emph{training-free}. \checkmark/\xmark/\pmark${=}$yes/no/partial.}
\label{tab:related}
\vspace{0.4em}\renewcommand{\arraystretch}{1.12}\setlength{\tabcolsep}{5pt}\small
\begin{tabular}{@{}l *{6}{c} @{}}
\toprule
 & \multicolumn{1}{c}{\rothdr{repr.\ present}}
 & \multicolumn{1}{c}{\rothdr{behavioral\\grounding test}}
 & \multicolumn{1}{c}{\rothdr{sign-\\inversion}}
 & \multicolumn{1}{c}{\rothdr{3-way\\taxonomy}}
 & \multicolumn{1}{c}{\rothdr{correction\\spectrum}}
 & \multicolumn{1}{c}{\rothdr{training-\\free}} \\
\midrule
Beyond Semantics \citep{beyondsemantics2025}          & \cmark & \xmark & \xmark & \xmark & \xmark & \xmark \\
3D Scene Topology \citep{scenetopology2026}            & \cmark & \xmark & \xmark & \xmark & \xmark & \xmark \\
Geometry of Failures \citep{geomfailures2026}          & \cmark & \xmark & \xmark & \xmark & \xmark & \cmark \\
Why Far Looks Up \citep{whyfarup2026}                  & \cmark & \pmark & \xmark & \xmark & \xmark & \cmark \\
Spatial IDs \citep{linearspatiotemporal2026}           & \cmark & \xmark & \xmark & \xmark & \xmark & \cmark \\
Actionability gap \citep{basu2026interp}               & \cmark & \pmark & \xmark & \xmark & \xmark & \cmark \\
\midrule
\rowcolor{groundedTint}
\textbf{Ours} & \cmark & \cmark & \cmark & \cmark & \cmark & \pmark \\
\bottomrule
\end{tabular}
\vspace{0.3em}\\
\footnotesize \pmark\ (behavioral): relate representation to behavior but not via a per-axis causal
vision-ablation control. \pmark\ (Ours, training-free): rotation is training-free on the clean-rotation
model; distributed inversions still need a trained edit.
\end{table}

\label{app:additional}

This appendix contains the supporting figures and tables referenced from the main text: the
layer-wise analyses (\S\ref{app:layers}), the full cross-model and external-benchmark results
(\S\ref{app:crossmodel}), the complete correction battery and population mechanism
(\S\ref{app:correction}), and robustness controls (\S\ref{app:robustness}).

\subsection{The dissociation, in detail}
\label{app:dissociation}
\begin{figure}[t]
\centering
\begin{tikzpicture}
\begin{axis}[
    width=0.92\linewidth, height=4.8cm,
    ybar=2pt, bar width=11pt,
    enlarge x limits=0.22,
    ymin=20, ymax=100,
    ytick={30,40,50,60,70,80,90,100},
    ymajorgrids, grid style={gray!18},
    ylabel={\small accuracy},
    symbolic x coords={Horizontal, Vertical, Depth},
    xtick=data, xticklabel style={font=\small},
    tick label style={font=\footnotesize},
    legend style={at={(0.5,1.04)}, anchor=south, legend columns=2,
                  font=\footnotesize, draw=neutralGray!50},
    nodes near coords, nodes near coords style={font=\tiny\bfseries, /pgf/number format/.cd, fixed, precision=1},
]
\addplot[draw=accentBlue, fill=accentBlue!25!white] coordinates {(Horizontal,84.9) (Vertical,94.1) (Depth,77.2)};
\addplot[draw=groundedTeal, fill=groundedTeal!28!white] coordinates {(Horizontal,81.3) (Vertical,58.5) (Depth,30.7)};
\draw[densely dashed, neutralGray!90!black, thick]
    (axis cs:Horizontal,50) -- (axis cs:Depth,50);
\node[font=\tiny, neutralGray!90!black] at (axis cs:Depth,46) {chance};
\legend{linear-probe peak decode, binary behavior}
\end{axis}
\end{tikzpicture}
\caption{\textbf{The decodability--behavior dissociation that invites a wrong conclusion.}
Peak probe-decode accuracy (blue) vs.\ actual binary behavior (teal). Horizontal is decoded
\emph{and} deployed (gap $+4$). Vertical is the \emph{most} decodable axis yet sits near
chance behaviorally (gap $+36$). Depth is decodable but behavior is \emph{below} chance
(gap $+47$, anti-correlated). Read alone, this looks like ``knowledge present, not deployed''
and motivates a recovery method --- the trap we spring in \S\ref{sec:arbiter}. Values shown as percentages.}
\label{fig:dissociation}
\end{figure}
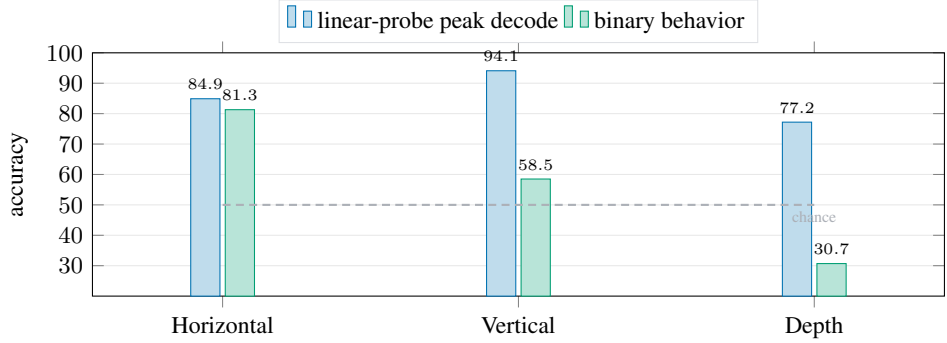

Figure~\ref{fig:dissociation} plots behavior against decodability for the anchor model, the visual
form of the main-text Table~\ref{tab:dissociation}.

\subsection{Layer-wise analyses}
\begin{table}[t]
\centering
\caption{\textbf{External-benchmark grounding (Tier-3).} Real vs.\ gray/mismatch balanced accuracy
(chance $50$; grounded-correct when real $\gg$ mismatch). The same models that invert depth on
ViewSpatial~\citep{viewspatial2025} (Table~\ref{tab:dissociation}) are grounded-correct on What'sUp
\citep{kamath2023whatsup} and 3DSRBench~\citep{ma20243dsrbench} depth, a depth-\emph{type} boundary; four
further models replicate the external grounding (Appendix~\ref{app:crossmodel}).}
\label{tab:external}
\vspace{0.3em}\renewcommand{\arraystretch}{1.12}\setlength{\tabcolsep}{6pt}\small
\begin{tabular}{l l c c c l}
\toprule
\textbf{benchmark} & \textbf{axis (task)} & \textbf{real} & \textbf{gray} & \textbf{mismatch} & \textbf{verdict} \\
\midrule
\multicolumn{6}{l}{\textit{Qwen2.5-VL-7B anchor}}\\
What'sUp-A & horizontal (left/right) & 99 & 51.5 & 51.5 & grounded \\
What'sUp-A & vertical (on/under)     & 99.5 & ---   & 47.6 & \textbf{grounded} \\
What'sUp-B & horizontal              & 100 & ---   & 47.1 & grounded \\
What'sUp-B & depth (front/behind)    & \textbf{99.5} & 50 & 48.5 & \textbf{grounded-correct} \\
3DSRBench  & vertical (height)       & 56.1 & 56.2 & 54.1 & prior \\
3DSRBench  & depth (closer-to-cam)   & \textbf{72.8} & 56.6 & 49.4 & \textbf{grounded-correct} \\
\bottomrule
\end{tabular}
\end{table}

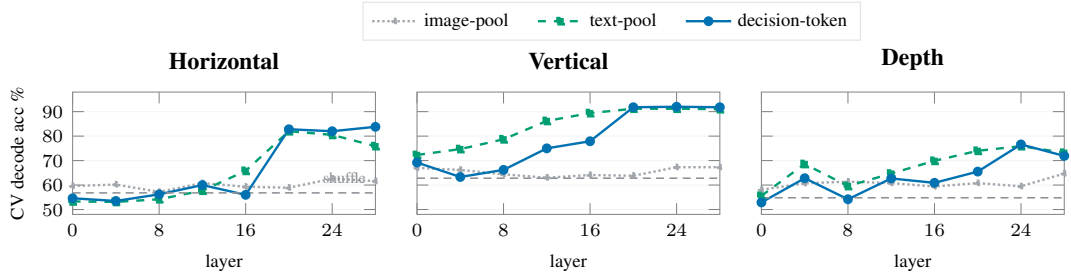
\begin{figure}[t]
\centering
\begin{tikzpicture}
\begin{groupplot}[
    group style={group size=3 by 1, horizontal sep=0.55cm, ylabels at=edge left},
    width=0.40\linewidth, height=3.2cm,
    xmin=0, xmax=28, xtick={0,8,16,24},
    ymin=48, ymax=98, ytick={50,60,70,80,90},
    tick label style={font=\scriptsize},
    xlabel={\scriptsize layer},
    ymajorgrids, grid style={neutralGray!18, very thin},
    axis line style={neutralGray!70!black},
    every axis plot/.append style={line width=0.9pt},
    title style={font=\footnotesize\bfseries, yshift=-1pt},
]
\nextgroupplot[title={Horizontal}, ylabel={\scriptsize CV decode acc \%},
    legend style={at={(1.78,1.42)}, anchor=south, legend columns=3,
                  font=\scriptsize, draw=neutralGray!50,
                  /tikz/every even column/.append style={column sep=0.3cm}}]
\addplot[neutralGray!85!black, mark=diamond*, mark size=1.2pt, densely dotted] coordinates {(0,59.6)(4,60.2)(8,57.2)(12,60.8)(16,59.3)(20,58.9)(24,62.6)(28,61.6)};
\addlegendentry{image-pool}
\addplot[groundedTeal, mark=square*, mark size=1.2pt, dashed] coordinates {(0,53.3)(4,53.1)(8,54.2)(12,57.7)(16,65.8)(20,82.0)(24,80.5)(28,75.9)};
\addlegendentry{text-pool}
\addplot[accentBlue, mark=*, mark size=1.4pt] coordinates {(0,54.6)(4,53.5)(8,56.2)(12,60.0)(16,56.0)(20,82.8)(24,82.0)(28,83.8)};
\addlegendentry{decision-token}
\addplot[neutralGray!75!black, densely dashed, line width=0.6pt, forget plot] coordinates {(0,56.8)(28,56.8)};
\node[anchor=south east, font=\tiny, neutralGray!75!black] at (axis cs:28,56.8) {shuffle};
\nextgroupplot[title={Vertical}, yticklabels={,,,,}]
\addplot[neutralGray!85!black, mark=diamond*, mark size=1.2pt, densely dotted] coordinates {(0,67.0)(4,66.2)(8,64.4)(12,63.0)(16,64.1)(20,63.8)(24,67.3)(28,67.3)};
\addplot[groundedTeal, mark=square*, mark size=1.2pt, dashed] coordinates {(0,72.3)(4,74.7)(8,78.7)(12,86.2)(16,89.4)(20,91.2)(24,91.2)(28,91.0)};
\addplot[accentBlue, mark=*, mark size=1.4pt] coordinates {(0,69.2)(4,63.3)(8,66.2)(12,75.0)(16,77.9)(20,91.8)(24,92.0)(28,91.8)};
\addplot[neutralGray!75!black, densely dashed, line width=0.6pt, forget plot] coordinates {(0,62.8)(28,62.8)};
\nextgroupplot[title={Depth}, yticklabels={,,,,}]
\addplot[neutralGray!85!black, mark=diamond*, mark size=1.2pt, densely dotted] coordinates {(0,58.1)(4,60.7)(8,61.4)(12,60.8)(16,59.4)(20,60.8)(24,59.5)(28,64.7)};
\addplot[groundedTeal, mark=square*, mark size=1.2pt, dashed] coordinates {(0,55.6)(4,68.6)(8,59.6)(12,64.7)(16,69.9)(20,74.0)(24,75.9)(28,73.2)};
\addplot[accentBlue, mark=*, mark size=1.4pt] coordinates {(0,52.9)(4,62.8)(8,54.2)(12,62.7)(16,60.9)(20,65.5)(24,76.6)(28,71.9)};
\addplot[neutralGray!75!black, densely dashed, line width=0.6pt, forget plot] coordinates {(0,54.8)(28,54.8)};
\end{groupplot}
\end{tikzpicture}
\caption{\textbf{The axis signal is assembled LM-side, not in raw image tokens} (Qwen2.5-VL-7B,
five-fold CV probe). Per-layer decode accuracy of the within-axis pole from three loci:
\textcolor{accentBlue}{decision token}, mean-pooled \textcolor{groundedTeal!70!black}{text tokens}, and
mean-pooled image tokens (grey). On all three axes the \emph{image-pool} trace stays flat near its
shuffle floor (dashed) while the decision/text traces rise sharply at L16--24: the direction is composed
in the residual stream LM-side, not read off image patches. Percentages.}
\label{fig:loci}
\end{figure}
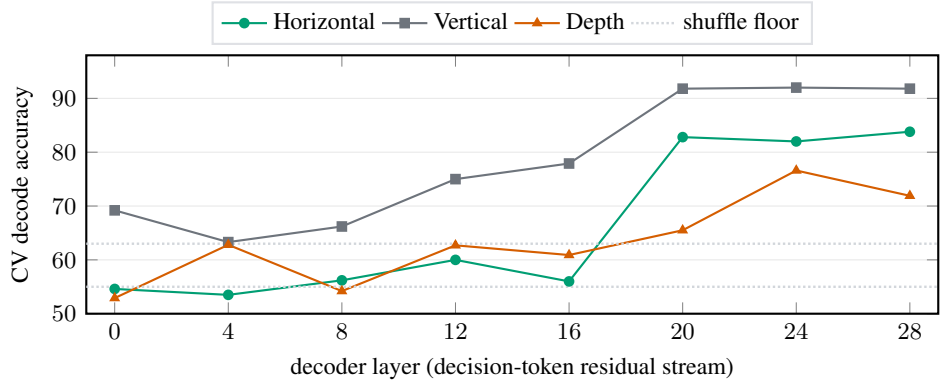
\begin{figure}[t]
\centering
\begin{tikzpicture}
\begin{axis}[
    width=0.92\linewidth, height=5.0cm,
    xmin=-1, xmax=29, ymin=50, ymax=98,
    xtick={0,4,8,12,16,20,24,28},
    ytick={50,60,70,80,90},
    xlabel={\small decoder layer (decision-token residual stream)},
    ylabel={\small CV decode accuracy},
    tick label style={font=\footnotesize},
    ymajorgrids, grid style={gray!18},
    legend style={at={(0.5,1.04)}, anchor=south, legend columns=4,
                  font=\footnotesize, draw=neutralGray!50},
    legend cell align=left,
    mark size=1.6pt, thick,
]
\addplot[groundedTeal, mark=*] coordinates {(0,54.6)(4,53.5)(8,56.2)(12,60)(16,56)(20,82.8)(24,82)(28,83.8)};
\addplot[priorSlate, mark=square*] coordinates {(0,69.2)(4,63.3)(8,66.2)(12,75)(16,77.9)(20,91.8)(24,92)(28,91.8)};
\addplot[invertedVerm, mark=triangle*] coordinates {(0,52.9)(4,62.8)(8,54.2)(12,62.7)(16,60.9)(20,65.5)(24,76.6)(28,71.9)};
\addplot[neutralGray!70, densely dotted, no marks, line width=0.9pt] coordinates {(-1,63)(29,63)};
\addplot[neutralGray!70, densely dotted, no marks, line width=0.9pt] coordinates {(-1,55)(29,55)};
\legend{Horizontal, Vertical, Depth, shuffle floor}
\end{axis}
\end{tikzpicture}
\caption{\textbf{All three axes are linearly decodable; vertical most of all.} Five-fold
cross-validated logistic-probe accuracy for the two within-axis poles, read from the
decision-token residual stream at each layer (PCA-50 features). Decoding rises sharply at
L16--24 for every axis. Vertical peaks \emph{highest} ($94$ at L23), above horizontal
($85$); depth reaches $77$. The dotted band is the per-axis shuffled-label floor
($55$--$63$). On a probe-only reading, all three axes ``contain'' the answer --- a
conclusion Fig.~\ref{fig:thesis} overturns for vertical. Values shown as percentages.}
\label{fig:decode_layers}
\end{figure}

\label{app:layers}
Complementing the main-text locus figure (Fig.~\ref{fig:loci}), the per-layer decision-token decode
peaks at L20--24 (Fig.~\ref{fig:decode_layers}), and the causal locus coincides with this decode onset
(Fig.~\ref{fig:s4_layers}). On the second model,
the inverted depth channel is causal, sharply localized, and one-sided across the population
(Fig.~\ref{fig:locality2}), with a layer-wise decode/controllability profile shown in
Fig.~\ref{fig:layerwise}.

\begin{figure}[t]
\centering
\begin{tikzpicture}
\begin{groupplot}[
    group style={group size=2 by 1, horizontal sep=1.4cm},
    width=0.50\linewidth, height=4.8cm,
    xmin=11, xmax=27, xtick={12,16,20,24},
    tick label style={font=\scriptsize},
    ymajorgrids, grid style={gray!18},
    every axis plot/.append style={thick, mark size=1.4pt},
    legend style={font=\scriptsize, draw=neutralGray!50, at={(0.03,0.97)}, anchor=north west},
    legend cell align=left,
]
\nextgroupplot[ylabel={\small recover $\Delta$ vs baseline (\%)}, ymin=-2, ymax=20,
               title={\small corrective injection}, title style={font=\footnotesize}]
\addplot[groundedTeal, mark=*] coordinates {(12,0)(16,0.7)(18,0.7)(20,7.3)(21,12)(22,12.7)(23,12)(24,10.7)(26,2.7)};
\addplot[priorSlate, mark=square*] coordinates {(12,0)(16,0)(18,0)(20,3.3)(21,16)(22,16)(23,8.7)(24,8)(26,0.7)};
\addplot[invertedVerm, mark=triangle*] coordinates {(12,0)(16,0.7)(18,0.7)(20,7.3)(21,13.3)(22,14.7)(23,12)(24,11.3)(26,12.7)};
\legend{H, V, D}
\nextgroupplot[ylabel={\small ablate $\Delta$ vs baseline (\%)}, ymin=-9, ymax=7,
               title={\small project-out (necessity)}, title style={font=\footnotesize}]
\addplot[groundedTeal, mark=*] coordinates {(12,0)(16,0)(18,0)(20,2)(21,0.7)(22,2.7)(23,3.3)(24,1.3)(26,2.7)};
\addplot[priorSlate, mark=square*] coordinates {(12,0)(16,0)(18,-0.7)(20,-0.7)(21,-6.7)(22,-5.3)(23,-4)(24,-2)(26,0)};
\addplot[invertedVerm, mark=triangle*] coordinates {(12,0)(16,0)(18,0)(20,0.7)(21,5.3)(22,6)(23,6)(24,1.3)(26,2)};
\addplot[neutralGray!70, densely dotted, no marks] coordinates {(11,0)(27,0)};
\end{groupplot}
\node[font=\footnotesize] at (3.2,-1.55) {decoder layer};
\end{tikzpicture}
\caption{\textbf{The deployment locus is a distributed L20--24 path, and vertical necessity is
localized there (S4).} \emph{Left:} corrective-injection gain peaks at L21--22 for all axes, matching
the decode onset (Fig.~\ref{fig:loci}). \emph{Right:} project-out (necessity) dips \emph{negative}
for vertical at L21--24 (peak $-6.7$ at L21), whereas horizontal stays flat (redundant code) and
depth is positive (removing the inverted signal helps). A five-layer band beats any single layer
(Table~\ref{tab:band}). $n{=}150$ subsample, so deltas are subsample-level; the shape and locus are
robust. Values shown as percentages.}
\label{fig:s4_layers}
\end{figure}
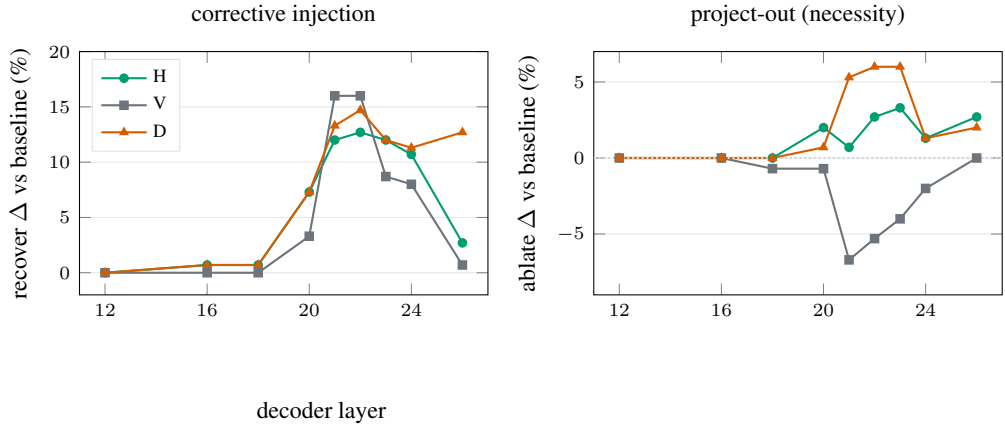
\begin{figure}[t]
\centering
\begin{tikzpicture}
\begin{axis}[
    width=0.84\linewidth, height=4.8cm,
    xmin=10, xmax=38,
    ymin=0, ymax=108,
    xtick={12,16,20,24,28,32,36},
    ytick={0,25,50,75,100},
    xlabel={\small layer (Qwen3-VL-8B)},
    ylabel={\small \%},
    tick label style={font=\footnotesize},
    ymajorgrids, grid style={neutralGray!22, very thin},
    axis line style={neutralGray!70!black},
    legend style={at={(0.5,1.04)}, anchor=south, legend columns=2,
                  font=\footnotesize, draw=neutralGray!50},
    legend cell align=left,
    every axis plot/.append style={line width=1.1pt},
]
\addplot[accentBlue, mark=*, mark size=1.6pt] coordinates {
    (12,55)(16,58)(20,68)(24,84)(26,88)(30,83)(33,82)(36,84)};
\addlegendentry{probe decode acc}
\addplot[invertedVerm, mark=triangle*, mark size=1.9pt] coordinates {
    (12,22)(16,22)(20,23)(24,98.7)(27,39)(30,35.1)(33,35.1)(36,36.4)};
\addlegendentry{causal controllability ($\beta{=}4$)}
\draw[neutralGray!65!black, densely dotted] (axis cs:24,0) -- (axis cs:24,98.7);
\node[anchor=south west, font=\tiny\bfseries, invertedVerm!85!black] at (axis cs:24.3,96) {L24 spike};
\node[anchor=south, font=\tiny, accentBlue!80!black] at (axis cs:28,89) {decode peak L26};
\node[anchor=north west, font=\tiny, neutralGray!75!black] at (axis cs:11,20) {flat, uncontrollable early};
\end{axis}
\end{tikzpicture}
\caption{\textbf{Layer-wise decode and causal controllability (Qwen3-VL-8B depth).} Probe-decoding
accuracy (blue) climbs from chance in the early layers to a peak at L26, while one-sided causal
controllability (vermillion, $\beta{=}4$) is flat and near-zero early, then \emph{spikes} to $98.7\%$
at L24 before falling back to $\sim$35--39\% in the later layers. The narrow controllability spike at
the decode-onset region---strongly asymmetric, recovering correct depth from a single-sign steer---is
the clean causal locus of the inversion; the best \emph{trained} edit instead prefers the decode-peak
L30 ($+57.8$ vs.\ $+38.3$ at L24, \S\ref{sec:method}). Early decode values ($<$L20) are approximate;
L20--L36 points and the controllability curve are measured. Values shown as percentages.}
\label{fig:layerwise}
\end{figure}
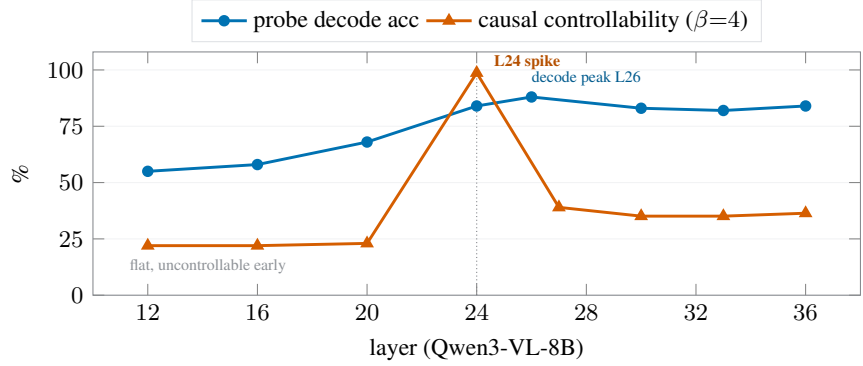

\subsection{Intervention geometry}
\label{app:intervention}
\begin{figure}[t]
\centering
\begin{tikzpicture}
\begin{axis}[
    width=0.84\linewidth, height=5.0cm,
    xmin=-4.4, xmax=4.4, ymin=25, ymax=90,
    xtick={-4,-2,-1,0,1,2,4},
    ytick={30,40,50,60,70,80},
    xlabel={\small steering strength $\alpha$ (std along $\hat{d}$, injected at decision token)},
    ylabel={\small $P(\text{$+$pole})$},
    tick label style={font=\footnotesize},
    ymajorgrids, grid style={gray!18},
    legend style={at={(0.5,1.04)}, anchor=south, legend columns=3,
                  font=\footnotesize, draw=neutralGray!50},
    legend cell align=left, mark size=1.8pt, thick,
]
\addplot[groundedTeal, mark=*] coordinates {(-4,28.8)(-2,39.8)(-1,45.2)(-0.5,47.7)(0,51.4)(0.5,54.6)(1,56.8)(2,62.5)(4,71.6)};
\addlegendentry{Horizontal (sign $+$)}
\addplot[priorSlate, mark=square*] coordinates {(-4,55.1)(-2,64.1)(-1,67.8)(-0.5,69.4)(0,71.8)(0.5,74.2)(1,75.8)(2,81.4)(4,85.1)};
\addlegendentry{Vertical (sign $+$)}
\addplot[invertedVerm, mark=triangle*] coordinates {(-4,56.9)(-2,51)(-1,50.3)(-0.5,47.7)(0,45.8)(0.5,41.8)(1,37.9)(2,33.3)(4,28.1)};
\addlegendentry{Depth (sign $-$)}
\addplot[neutralGray!90!black, densely dashed, no marks, line width=0.8pt] coordinates {(-4.4,50)(4.4,50)};
\end{axis}
\end{tikzpicture}
\caption{\textbf{Causal control is clean and monotone on all three axes (S3).} Probability of the
positive pole as the per-axis diff-of-means direction is added at the decision token (full $n{=}1047$,
Qwen2.5-VL-7B). Horizontal and vertical increase with $\alpha$ (auto-detected readout sign $+1$);
depth \emph{decreases} (sign $-1$), the steering signature of an inverted readout. Each axis is
causally steerable---which, like decodability (Fig.~\ref{fig:decode_layers}), is necessary but not
sufficient for grounding: the arbiter (\S\ref{sec:arbiter}) shows the vertical lever moves a prior. Values shown as percentages.}
\label{fig:causal_sweep}
\end{figure}
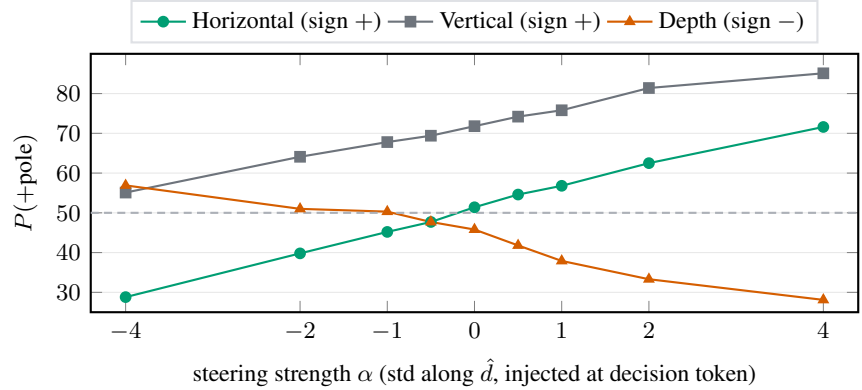

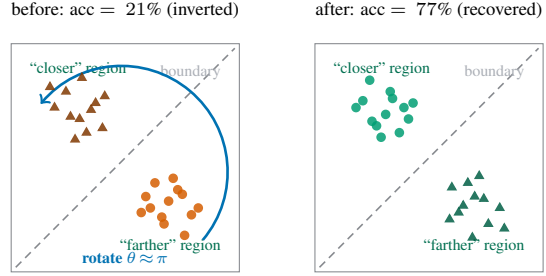
\begin{figure}[t]
\centering
\begin{tikzpicture}
\begin{groupplot}[
    group style={group size=2 by 1, horizontal sep=1.0cm},
    width=4.6cm, height=4.6cm,
    xmin=-2.5, xmax=2.5, ymin=-2.5, ymax=2.5,
    xtick=\empty, ytick=\empty,
    axis line style={neutralGray!70!black},
    enlargelimits=false,
    every axis plot/.append style={line width=0.6pt},
]

\nextgroupplot[title={\scriptsize before: acc $=\,21$\% (inverted)}]
\addplot[neutralGray!70!black, densely dashed, no marks] coordinates {(-2.4,-2.4)(2.4,2.4)};
\node[anchor=south east, font=\tiny, neutralGray!90!black] at (axis cs:2.3,1.5) {boundary};
\node[anchor=north west, font=\tiny, groundedTeal!70!black] at (axis cs:-2.3,2.3) {``closer'' region};
\node[anchor=south east, font=\tiny, groundedTeal!70!black] at (axis cs:2.3,-2.3) {``farther'' region};
\addplot[only marks, mark=*, mark size=1.5pt, invertedVerm, opacity=0.85] coordinates {
    (0.55,-1.10)(0.85,-1.45)(1.15,-0.70)(0.65,-0.55)(1.45,-1.25)
    (0.95,-0.95)(1.30,-1.70)(0.45,-0.85)(1.60,-1.10)(0.80,-1.30)
    (1.05,-0.45)(0.35,-1.20)(1.25,-0.80)
};
\addplot[only marks, mark=triangle*, mark size=1.7pt, invertedVerm!60!black, opacity=0.85] coordinates {
    (-0.65,1.20)(-1.00,0.85)(-1.30,1.45)(-0.50,0.65)(-1.55,1.10)
    (-0.85,0.55)(-1.20,0.90)(-0.45,1.35)(-1.65,1.55)(-0.75,1.05)
    (-1.10,0.40)(-0.95,1.75)
};
\draw[->, accentBlue, thick, line width=1.0pt]
    (axis cs:1.7,-1.8) arc[start angle=-40, end angle=140, radius=1.4cm];
\node[anchor=center, font=\tiny\bfseries, accentBlue] at (axis cs:0,-2.2) {rotate $\theta\!\approx\!\pi$};

\nextgroupplot[title={\scriptsize after: acc $=\,77$\% (recovered)}]
\addplot[neutralGray!70!black, densely dashed, no marks] coordinates {(-2.4,-2.4)(2.4,2.4)};
\node[anchor=south east, font=\tiny, neutralGray!90!black] at (axis cs:2.3,1.5) {boundary};
\node[anchor=north west, font=\tiny, groundedTeal!70!black] at (axis cs:-2.3,2.3) {``closer'' region};
\node[anchor=south east, font=\tiny, groundedTeal!70!black] at (axis cs:2.3,-2.3) {``farther'' region};
\addplot[only marks, mark=*, mark size=1.5pt, groundedTeal, opacity=0.85] coordinates {
    (-0.55,1.10)(-0.85,1.45)(-1.15,0.70)(-0.65,0.55)(-1.45,1.25)
    (-0.95,0.95)(-1.30,1.70)(-0.45,0.85)(-1.60,1.10)(-0.80,1.30)
    (-1.05,0.45)(-0.35,1.20)(-1.25,0.80)
};
\addplot[only marks, mark=triangle*, mark size=1.7pt, groundedTeal!60!black, opacity=0.85] coordinates {
    (0.65,-1.20)(1.00,-0.85)(1.30,-1.45)(0.50,-0.65)(1.55,-1.10)
    (0.85,-0.55)(1.20,-0.90)(0.45,-1.35)(1.65,-1.55)(0.75,-1.05)
    (1.10,-0.40)(0.95,-1.75)
};
\end{groupplot}
\end{tikzpicture}
\caption{\textbf{The depth inversion is a clean low-dimensional rotation in Qwen3-VL-8B.} Schematic
of the depth readout plane. \textbf{Left}: the model's readout places true-``closer'' items on the
``farther'' side of the decision boundary (and vice-versa)---a systematic sign inversion of an
otherwise well-separated representation (accuracy $21.4$, well below chance). \textbf{Right}: a
single training-free, norm-preserving rotation by $\theta{\approx}\pi$ re-aligns the clusters with
the correct labels (accuracy $77.1$, matching a trained low-rank edit). When the inversion is a
clean rotation, the un-inversion is too---this is the geometry that lets a zero-training fix
re-deploy the axis (\S\ref{sec:method}). Values shown as percentages.}
\label{fig:rotation}
\end{figure}

The causal sweep (Fig.~\ref{fig:causal_sweep}) and the diff-of-means projection
(Fig.~\ref{fig:projection}) establish causal control; the un-inversion is a near-$\pi$ rotation of the
readout (Fig.~\ref{fig:rotation}).

\subsection{Full cross-model and external results}
\label{app:crossmodel}
\begin{figure}[t]
\centering
\begin{tikzpicture}
\begin{axis}[
    width=\linewidth, height=5.6cm,
    ybar=0.6pt, bar width=4.0pt,
    enlarge x limits=0.055,
    ymin=20, ymax=100,
    ytick={30,40,50,60,70,80,90,100},
    ymajorgrids, grid style={gray!18},
    ylabel={\small real balanced accuracy},
    symbolic x coords={IV3-2B, IV3-8B, IV3-14B, Qw2.5-3B, Qw2.5-7B, LLaVA-7B, Qw3-8B, Pixtral-12B, SpaceOm, SmolVLM-2B},
    xtick=data,
    xticklabel style={font=\scriptsize, rotate=30, anchor=east},
    tick label style={font=\footnotesize},
    legend style={at={(0.5,1.04)}, anchor=south, legend columns=3,
                  font=\footnotesize, draw=neutralGray!50},
    legend cell align=left,
]
\addplot[draw=groundedTeal, fill=groundedTeal!28!white] coordinates {(IV3-2B,65.8)(IV3-8B,86)(IV3-14B,88)(Qw2.5-3B,70.7)(Qw2.5-7B,82.6)(LLaVA-7B,75.3)(Qw3-8B,87.7)(Pixtral-12B,70.8)(SpaceOm,75.6)(SmolVLM-2B,47.4)};
\addplot[draw=priorSlate, fill=priorSlate!28!white] coordinates {(IV3-2B,57.5)(IV3-8B,89.5)(IV3-14B,90)(Qw2.5-3B,65.9)(Qw2.5-7B,69.8)(LLaVA-7B,66.6)(Qw3-8B,94.6)(Pixtral-12B,74.5)(SpaceOm,65.3)(SmolVLM-2B,57.4)};
\addplot[draw=invertedVerm, fill=invertedVerm!28!white] coordinates {(IV3-2B,38.1)(IV3-8B,30.9)(IV3-14B,28.2)(Qw2.5-3B,47.9)(Qw2.5-7B,31.6)(LLaVA-7B,37.8)(Qw3-8B,22.4)(Pixtral-12B,42.2)(SpaceOm,47.3)(SmolVLM-2B,47.8)};
\draw[densely dashed, neutralGray!90!black, thick]
 (axis cs:IV3-2B,50) -- (axis cs:SmolVLM-2B,50);
\node[anchor=south east, font=\tiny, neutralGray!90!black, fill=white, inner sep=0.6pt]
 at (axis cs:SmolVLM-2B,50) {chance};
\legend{Horizontal, Vertical (prior), Depth}
\end{axis}
\end{tikzpicture}
\caption{\textbf{The three-regime taxonomy is architecture-general (10 of 14 models shown; full set in
Table~\ref{tab:xmodel}).} Real balanced accuracy per axis. Horizontal (teal) is high and
vision-dependent in every \emph{capable} model; vertical (slate) is above chance but prior-driven
everywhere (real $\approx$ gray $\approx$ mismatch; Table~\ref{tab:xmodel}) and is the most
probe-decodable axis; depth (vermillion) is at or below chance in all and inverted in the larger models.
The lone all-axis exception is SmolVLM-2.2B (horizontal $\approx$ chance), a capability floor rather
than a counterexample. The newest model (Qwen3-VL-8B) is the most depth-inverted; the non-Qwen
Pixtral-12B (Mistral) still shows the full taxonomy. Values shown as percentages.}
\label{fig:xmodel}
\end{figure}
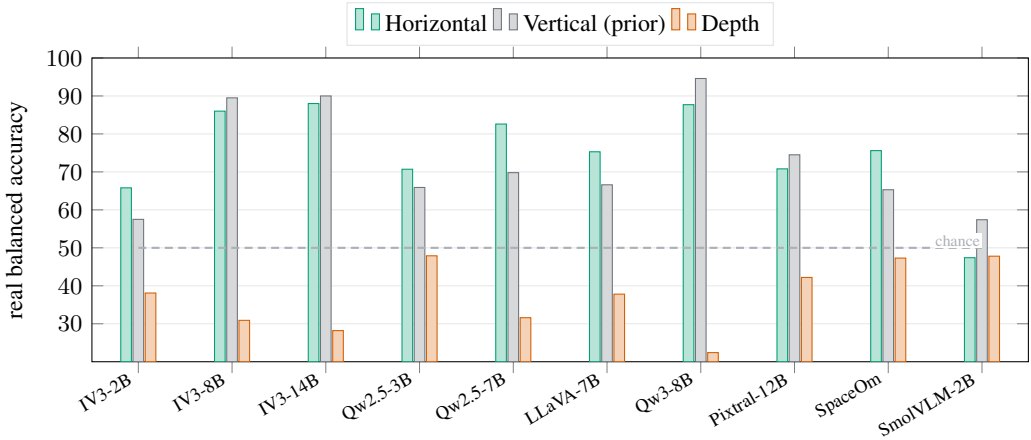

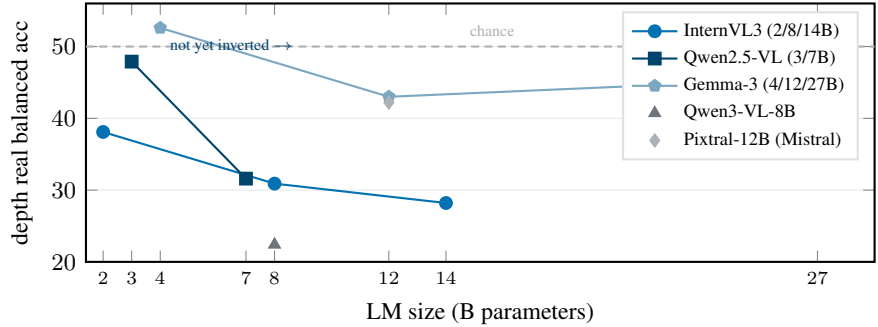
\begin{figure}[t]
\centering
\begin{tikzpicture}
\begin{axis}[
    width=0.86\linewidth, height=5.0cm,
    xmin=1.4, xmax=29, ymin=20, ymax=56,
    xtick={2,3,4,7,8,12,14,27}, xticklabel style={font=\scriptsize},
    ytick={20,30,40,50},
    xlabel={\small LM size (B parameters)},
    ylabel={\small depth real balanced acc},
    tick label style={font=\footnotesize},
    ymajorgrids, grid style={gray!18},
    legend style={at={(0.98,0.97)}, anchor=north east, font=\scriptsize, draw=neutralGray!50},
    legend cell align=left, mark size=2.2pt, thick,
]
\addplot[accentBlue, mark=*] coordinates {(2,38.1)(8,30.9)(14,28.2)};
\addlegendentry{InternVL3 (2/8/14B)}
\addplot[accentBlue!60!black, mark=square*] coordinates {(3,47.9)(7,31.6)};
\addlegendentry{Qwen2.5-VL (3/7B)}
\addplot[accentBlue!35!neutralGray, mark=pentagon*] coordinates {(4,52.6)(12,43)(27,45.6)};
\addlegendentry{Gemma-3 (4/12/27B)}
\addplot[priorSlate, only marks, mark=triangle*] coordinates {(8,22.4)};
\addlegendentry{Qwen3-VL-8B}
\addplot[neutralGray!90!black, only marks, mark=diamond*] coordinates {(12,42.2)};
\addlegendentry{Pixtral-12B (Mistral)}
\addplot[neutralGray!90!black, densely dashed, no marks, line width=0.8pt] coordinates {(1.4,50)(29,50)};
\node[anchor=south west, font=\tiny, neutralGray!90!black] at (axis cs:14.5,50) {chance};
\node[font=\tiny, accentBlue!60!black, anchor=west] at (axis cs:4.0,50) {not yet inverted $\rightarrow$};
\end{axis}
\end{tikzpicture}
\caption{\textbf{Depth inversion emerges with LM scale within families, but the size--severity relation
is family-specific and not strictly monotone.} InternVL3 ($2B\,38.1\!\to\!8B\,30.9\!\to\!14B\,28.2$)
and Qwen2.5-VL ($3B\,47.9\!\to\!7B\,31.6$) deepen monotonically with size, whereas Gemma-3
($4B\,52.6\!\to\!12B\,43\!\to\!27B\,45.6$) emerges then \emph{partially recovers} at 27B---a clean
within-family non-monotonicity. \emph{Onset} differs by family (InternVL3 already inverted at 2B;
Qwen2.5-VL-3B and Gemma-4B not yet), and across families severity is not size-monotonic (Pixtral-12B
milder than Qwen2.5-VL-7B; newest Qwen3-VL-8B the most inverted). The honest summary is ``inversion
emerges with scale within families; degree and monotonicity are family-specific.'' Values shown as percentages.}
\label{fig:depth_scale}
\end{figure}

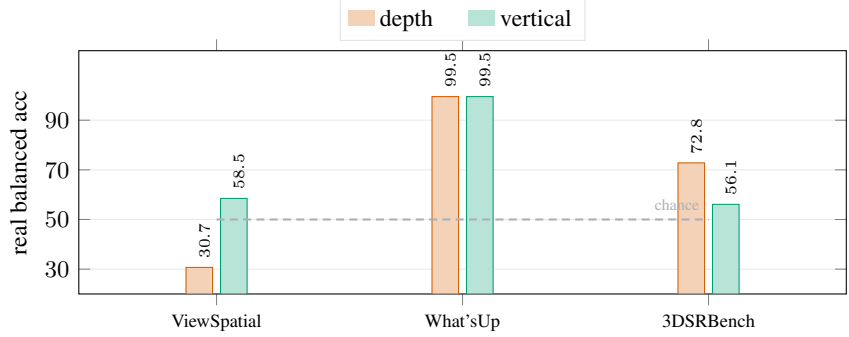
\begin{figure}[t]
\centering
\begin{tikzpicture}
\begin{axis}[
    width=0.84\linewidth, height=4.8cm,
    ybar=3pt, bar width=10pt,
    enlarge x limits=0.28,
    ymin=20, ymax=118,
    ytick={30,50,70,90},
    ymajorgrids, grid style={gray!18},
    ylabel={\small real balanced acc},
    symbolic x coords={ViewSpatial, WhatsUp, DSRBench},
    xtick=data,
    xticklabels={ViewSpatial, What'sUp, 3DSRBench},
    xticklabel style={font=\scriptsize},
    tick label style={font=\footnotesize},
    legend style={at={(0.5,1.04)}, anchor=south, legend columns=2,
                  font=\footnotesize, draw=neutralGray!50,
                  /tikz/every even column/.append style={column sep=0.4cm}},
    legend image code/.code={
      \draw[#1, draw=none] (0cm,-0.08cm) rectangle (0.35cm,0.13cm);
    },
    nodes near coords, nodes near coords style={font=\tiny\bfseries, rotate=90, anchor=west, yshift=-2pt, /pgf/number format/.cd, fixed, precision=1},
    every node near coord/.append style={/tikz/font=\tiny\bfseries},
]
\addplot[draw=invertedVerm, fill=invertedVerm!28!white] coordinates {(ViewSpatial,30.7) (WhatsUp,99.5) (DSRBench,72.8)};
\addlegendentry{depth}
\addplot[draw=groundedTeal, fill=groundedTeal!28!white] coordinates {(ViewSpatial,58.5) (WhatsUp,99.5) (DSRBench,56.1)};
\addlegendentry{vertical}
\draw[densely dashed, neutralGray!90!black, thick]
 (axis cs:ViewSpatial,50) -- (axis cs:DSRBench,50);
\node[anchor=south east, font=\tiny, neutralGray!90!black] at (axis cs:DSRBench,50) {chance};
\end{axis}
\end{tikzpicture}
\caption{\textbf{Grounding is task-type-specific: the depth inversion is ViewSpatial-camera-egocentric,
not a general depth failure} (Qwen2.5-VL-7B; depth on What'sUp-B front/behind and 3DSRBench
closer-to-camera, vertical on What'sUp-A on/under and 3DSRBench height). The \emph{same} model that
inverts depth on ViewSpatial's camera-relative front/back ($30.7$) is grounded-correct on near-field
front/behind ($99.5$) and metric closer-to-camera ($72.8$); mismatch controls confirm these are
vision-driven ($48.5$/$49.4$). Vertical is grounded on simple tabletop on/under ($99.5$) but a prior
on both ViewSpatial and 3DSRBench real-scene height. Horizontal is grounded everywhere externally
($99$--$100$). This pre-empts the ``ViewSpatial artifact'' objection: the inversion is a specific,
reproducible property of camera-egocentric front/back framing, not a broken benchmark. Values shown as percentages.}
\label{fig:external}
\end{figure}

Table~\ref{tab:xmodel} is the complete fourteen-model roster behind the matrix figure
(Fig.~\ref{fig:cardgrid}); Fig.~\ref{fig:xmodel} summarizes per-axis behavior and
Fig.~\ref{fig:depth_scale} the within-family scale ladders. On external benchmarks the inversion
becomes a task-type boundary (Fig.~\ref{fig:external}, Table~\ref{tab:external}): the same models that
invert ViewSpatial camera-egocentric front/back are grounded-correct on near-field front/behind and
metric closer-to-camera depth.

\begin{table}[t]
\centering
\caption{\textbf{Cross-model grounding matrix: 14 VLMs, six LM families.} Real balanced accuracy per
axis (chance $50$); \colorbox{rowcatE}{H grounded-correct}, \colorbox{rowcatB}{V prior},
\colorbox{rowcatD}{D inverted}. The taxonomy holds in every \emph{capable} model, including the
non-Qwen Pixtral (Mistral) and Gemma-3 (SigLIP vision). Shaded rows are the three within-family scale
ladders. The probe still decodes vertical at $87$--$97$ and depth at $73$--$86$ in every
capable model, so decodability overstates usable knowledge universally. All values are balanced accuracy in \%.}
\label{tab:xmodel}
\vspace{0.3em}\renewcommand{\arraystretch}{1.08}\setlength{\tabcolsep}{5pt}\footnotesize
\begin{tabular}{l l c c c l}
\toprule
\textbf{model} & \textbf{LM family} & \textbf{H} & \textbf{V} & \textbf{D} & \textbf{note} \\
\midrule
\rowcolor{rowours!50}
InternVL3-2B~\citep{internvl3}   & Qwen2-based  & 65.8 & 57.5 & \textbf{38.1} & ladder: already inverted at 2B \\
\rowcolor{rowours!50}
InternVL3-8B   & Qwen2-7B     & 86 & 89.5 & 30.9 & ladder (monotone) \\
\rowcolor{rowours!50}
InternVL3-14B  & Qwen2.5-14B  & 88 & 90 & \textbf{28.2} & ladder: deepest \\
\rowcolor{rowcatC!50}
Gemma-3-4B~\citep{gemma3}     & Gemma/SigLIP & \textit{49.5} & --- & 52.6 & ladder: H \& D not yet (small) \\
\rowcolor{rowcatC!50}
Gemma-3-12B    & Gemma/SigLIP & 67.3 & 75.9 & \textbf{43} & ladder: deepest \\
\rowcolor{rowcatC!50}
Gemma-3-27B    & Gemma/SigLIP & 69.8 & 78.4 & 45.6 & ladder: \emph{partial recovery} (non-mono.) \\
Qwen2.5-VL-3B~\citep{qwen2vl}  & Qwen2.5      & 70.7 & 65.9 & \textbf{47.9} & depth not yet inverted \\
Qwen2.5-VL-7B  & Qwen2.5      & 82.6 & 69.8 & 31.6 & primary (deep mechanism) \\
LLaVA-OV-7B~\citep{llava_onevision}    & Qwen2-7B     & 75.3 & 66.6 & 37.8 & $\approx$ InternVL3-8B (shared LM) \\
Qwen3-VL-8B~\citep{qwen3vl}    & Qwen3        & 87.7 & 94.6 & \textbf{22.4} & newest; most inverted \\
Pixtral-12B~\citep{pixtral2024}    & Mistral      & 70.8 & 74.5 & 42.2 & \textbf{non-Qwen}; taxonomy holds \\
SpaceOm~\citep{spacevlms}        & Qwen2.5 (sp-FT) & 75.6 & 65.3 & 47.3 & spatial-FT did not fix V/D$^{\dagger}$ \\
SpaceQwen-3B~\citep{spacevlms}   & Qwen2.5 (sp-FT) & \textit{50.2} & 56.2 & \textit{49.9} & near-chance; not a donor$^{\dagger}$ \\
SmolVLM-2.2B~\citep{smolvlm2025}   & SmolLM       & \textit{47.4} & 57.4 & 47.8 & capability floor (even H $\approx$ chance) \\
\bottomrule
\end{tabular}
\vspace{0.2em}\\
\footnotesize Shaded = within-family scale ladders (InternVL3 and Qwen2.5 monotone; Gemma non-monotone,
27B partially recovers; Fig.~\ref{fig:depth_scale}). $^{\dagger}$The two spatial-fine-tuned models
(SpaceOm, SpaceQwen) are grounded-correct on no failing axis, so neither is a clean ``donor''; SpaceOm
is a thinking model whose first-token eval likely understates it. SmolVLM-2.2B is near-chance on all
axes, bounding ``horizontal universal'' to capable models.
\end{table}

\subsection{Correction battery, population, and second-model mechanism}
\label{app:correction}
Table~\ref{tab:method} is the full nine-correction battery behind the main-text
Fig.~\ref{fig:method}; Table~\ref{tab:population} is the eight-model population behind the main-text
forest plot (Fig.~\ref{fig:forest}); Table~\ref{tab:model2} reproduces the deep mechanism on a second
model.
\begin{table}[t]
\centering
\caption{\textbf{The minimal correction that re-deploys depth diagnoses the inversion's geometry}
(test-acc lift over baseline; ViewSpatial~\citep{viewspatial2025} depth, $n{\approx}150$; baselines
qwen $26.2$ / qwen3-vl $21.4$ / pixtral $39.8$ / 3B $48.1$). Across nine corrections, a
\emph{training-free} norm-preserving rotation re-deploys the axis where the inversion is a clean
low-dimensional flip (Qwen3-VL-8B~\citep{qwen3vl}), \emph{matching} a trained edit; a distributed or
decodable-but-not-installable inversion needs the trained edit; and the un-inverted 3B (not inverted)
admits no correction. Lifts in percentage points over baseline.}
\label{tab:method}
\vspace{0.3em}\renewcommand{\arraystretch}{1.1}\setlength{\tabcolsep}{6pt}\small
\begin{tabular}{l c c c c}
\toprule
\textbf{correction} & \textbf{qwen-7B} & \textbf{qwen3-vl-8B} & \textbf{pixtral-12B} & \textbf{3B} \\
\midrule
recalibrate ($1$-D scalar)      & $+11.2$ & $+0$ & $+1$ & $+0$ \\
logit-lens reflect-at-onset     & n/a      & $+0$ & $+0.7$ & $-2.7$ \\
\rowcolor{rowcatE}
angular rotate ($\theta{\approx}\pi$)~\citep{angular2025} & $+8.8$ & $\mathbf{+55.7}$ & $+0.7$ & $+8.8$ \\
\rowcolor{rowcatE}
selective multi-layer rotate~\citep{angular2025}    & $+15.5$ & $\mathbf{+54.8}$ & $+2.4$ & $+1.9$ \\
SEA~\citep{sea2024} / SADI~\citep{sadi2024} edits     & $+10$ & $+6$ & $+1$ & $+1.4$ \\
ITI per-head reflect~\citep{iti2023}            & $+5.5$ & $+1.4$ & $+0$ & $+0.2$ \\
SAE feature steer$^{\ddagger}$~\citep{pach2025saevlm}  & $+2.9$ & $+1.2$ & $+0.2$ & $-1.2$ \\
\rowcolor{rowcatD}
LoRA~\citep{lora2021} / ReFT~\citep{wu2024reft} (trained)                  & $\mathbf{+42.9}$ & $\mathbf{+57.1}$ & $\mathbf{+32.5}$ & $-2.6$ \\
\midrule
\textbf{card-C rung} & reflect (1-D) & \textbf{rotate (2-D)} & none (trained) & none (not inv.) \\
\textbf{inversion geometry} & distributed & \textbf{clean rotation} & not-installable & flat (not inv.) \\
\bottomrule
\end{tabular}
\vspace{0.2em}\\
\footnotesize Shaded: \colorbox{rowcatE}{training-free rotation wins}, \colorbox{rowcatD}{trained edit
needed}. On Qwen3-VL-8B the training-free rotation ($+55.7$, $21.4\!\to\!77.1$, test $>$ val: not
overfit) matches trained LoRA ($+57.1$). $^{\ddagger}$The SAE feature diagnostic is exploratory and is not used as evidence. The training-free$\approx$trained match on Qwen3-VL-8B holds over three
seeds (\S\ref{sec:method}), and the population picture is the decode$\neq$deploy forest of
\S\ref{sec:method}. The \emph{card-C rung} is the minimal-complexity fixed correction
within margin of the best lift (a per-model effective-dimensionality fingerprint); the four models
shown are clear cases; borderline models (e.g.\ InternVL3-8B/14B) are reported separately.
\end{table}

\begin{table}[t]
\centering
\caption{\textbf{The inversion replicates across families; the minimal correction is per-model.}
Depth, $n{\approx}150$, single seed; lifts over the no-edit baseline. Four newly added models
(top) plus the three correction-battery models (bottom) all show inverted deployment
($a_{\text{decode}}{>}0$, $a_{\text{deploy}}{<}0$); the training-free angular rotation that
re-deploys depth on Qwen3-VL-8B~\citep{qwen3vl} does \emph{not} generalize cleanly, confirming that the rotation tie
is model-specific and that the population claim is the universal inversion plus a per-model
correction-complexity spectrum (\S\ref{sec:method}). Lift values shown as percentages.}
\label{tab:population}
\vspace{0.3em}\renewcommand{\arraystretch}{1.12}\setlength{\tabcolsep}{5pt}\small
\begin{tabular}{l l c c c c c l}
\toprule
\textbf{model} & \textbf{family} & \textbf{base} & $a_{\text{dec}}$ & $a_{\text{dep}}$ & \textbf{rotate}$^{*}$ & \textbf{recalib} & \textbf{correctability} \\
\midrule
InternVL3-2B~\citep{internvl3}   & InternVL3 & 38.3 & $+.74$ & $-.11$ & $-2.1$  & $+1.4$  & suppress-only \\
InternVL3-8B   & InternVL3 & 26.4 & $+.30$ & $-.07$ & $+8.1$  & $+2.6$  & mild rotation \\
InternVL3-14B  & InternVL3 & 31.4 & $+.50$ & $-.19$ & $-1.0$  & $\mathbf{+6.2}$ & recalib/reflect \\
LLaVA-OV-7B~\citep{llava_onevision}    & LLaVA     & 36.2 & $+.17$ & $-.13$ & $\mathbf{+13.8}$ & $+7.9$ & moderate rotation \\
\midrule
\rowcolor{groundedTint}
Qwen3-VL-8B    & Qwen3     & 21.4 & $+.08$ & $-.15$ & $\mathbf{+55.7}$ & $+6.0$ & clean rotation $=$ trained \\
Qwen2.5-VL-7B  & Qwen2.5   & 26.2 & $+.23$ & $-.08$ & $+15.5$ & $+11.2$ & distributed (trained) \\
Pixtral-12B~\citep{pixtral2024}    & Mistral   & 39.8 & $+1.59$& $-.25$ & $+2.4$  & $+1.0$  & not-installable (trained) \\
\midrule
Qwen2.5-VL-3B~\citep{qwen2vl}  & Qwen2.5   & 48.1 & $+.30$ & $+.01$ & $+8.8$  & $+0.0$  & \emph{aligned-flat (not inverted)} \\
\bottomrule
\end{tabular}
\vspace{0.2em}\\
\footnotesize $a_{\text{dec}}/a_{\text{dep}}$: signed steering slopes for probe-decodability and
behaviour (both $>0$ in the only non-inverted model). \textbf{7 of 8 models across 5 families are
inverted}; only the 3B scale-floor is not. rotate$^{*}$ is the best norm-preserving angular rotation
(training-free); recalib is the best 1-D scalar grid lift.
\end{table}

\begin{table}[t]
\centering
\caption{\textbf{The mechanism replicates on a second model (Qwen3-VL-8B~\citep{qwen3vl} depth, base $21.4\%$).}
The deep localization-and-correction story established on the Qwen2.5-VL-7B~\citep{qwen2vl} anchor reproduces
qualitatively: a trained rank-1 edit recovers the axis (matching the anchor's trained lift), and a
4-component ICA-subspace reflection also works---while the single probe direction and a closed-form
GEVD subspace both fail. The correctable channel is low-dimensional yet \emph{off} the probe axis.
Lift values shown as percentages.}
\label{tab:model2}
\vspace{0.3em}\renewcommand{\arraystretch}{1.12}\setlength{\tabcolsep}{6pt}\small
\begin{tabular}{l l c c l}
\toprule
\textbf{correction} & \textbf{best setting} & \textbf{test} & \textbf{lift} & \textbf{reading} \\
\midrule
\rowcolor{groundedTint}
ReFT~\citep{wu2024reft} (trained) & rank 1, L30 & $\mathbf{79.2}$ & $\mathbf{+57.8}$ & low-rank trainable (val${\approx}$test) \\
\rowcolor{groundedTint}
ICA-subspace reflect  & $k{=}4$, L30 & $\mathbf{81.0}$ & $\mathbf{+59.5}$ & multi-component, off probe axis \\
probe-direction reflect & $k{=}1{\dots}8$ & $21.4$ & $+0.0$ & single probe axis \emph{fails} \\
GEVD subspace (closed-form) & $k{=}1$, L30 & $21.7$ & $+0.2$ & closed-form basis \emph{fails} \\
\bottomrule
\end{tabular}
\vspace{0.2em}\\
\footnotesize The trained ReFT lift ($+57.8$) matches this model's correction-battery LoRA
($+57.1$, Table~\ref{tab:method}, LoRA~\citep{lora2021}). That ICA ($k{=}4$) succeeds where the single probe direction does
not echoes the cheap-lens vs.\ probe contrast and locates the readout error in a small,
non-probe-aligned subspace.
\end{table}

\subsection{Robustness controls and auxiliary analyses}
\label{app:robustness}
\begin{table}[t]
\centering
\caption{\textbf{Causal localization: multi-layer band, full $n$} (Qwen2.5-VL-7B~\citep{qwen2vl}, T16b).
Corrective-injection (\textbf{recover}, GT-signed) and project-out (\textbf{ablate}) at a 5-layer
band around each axis's decode peak. \emph{Recover} measures whether the readout \emph{can} use the
direction if it is written in (causal usability, not a deployable accuracy gain); \emph{ablate}
measures necessity. Deltas vs.\ baseline in parentheses. Accuracies in \%; deltas in percentage points.}
\label{tab:band}
\vspace{0.3em}\renewcommand{\arraystretch}{1.15}\setlength{\tabcolsep}{6pt}\small
\resizebox{\linewidth}{!}{%
\begin{tabular}{l c c c c l}
\toprule
\textbf{Axis} & \textbf{band} & \textbf{baseline} & \textbf{recover} & \textbf{ablate} & \textbf{reading} \\
\midrule
Horizontal & L19--23 & 82.4 & 98.1 ($+15.6$) & 82.2 ($-0.2$) & deployed; ablate flat (redundant) \\
Vertical   & L21--25 & 58.5 & 70.2 ($+11.7$) & \textbf{49.2} ($-9.3$) & recoverable \& necessary (ablate$\to$chance) \\
Depth      & L19--23 & 31.4 & 62.7 ($+31.4$) & 36.6 ($+5.2$) & sign-flip crosses chance; ablate helps \\
\bottomrule
\end{tabular}%
}
\vspace{0.2em}\\
\footnotesize Recover uses a GT-signed corrective injection: it shows the direction is causally
\emph{usable}, not that test accuracy improves. The arbiter (Table~\ref{tab:arbiter}) shows the
vertical lever operates on a prior rather than on vision.
\end{table}

\begin{table}[t]
\centering
\caption{\textbf{Decoding-optimal $\neq$ intervention-optimal} (Qwen2.5-VL-7B~\citep{qwen2vl}, T14, full $n$).
The probe direction (which maximizes decode, e.g.\ vertical $94$) steers $\sim$3$\times$ weaker than
diff-of-means and is not even necessary (vertical probe-ablate $57.7\approx$ baseline $58.5\approx$
random $58.6$). \textbf{bias} = $P(+\text{pole})$ span over the $\alpha$ sweep; \textbf{oracle} =
corrective-oracle peak; \textbf{ablate} = project-out (random in parentheses). Bias ranges are $P(+\text{pole})$ in \%; oracle/ablate in \% accuracy.}
\label{tab:probe_vs_diffmean}
\vspace{0.3em}\renewcommand{\arraystretch}{1.1}\setlength{\tabcolsep}{6pt}\small
\begin{tabular}{l l c c c}
\toprule
\textbf{Axis} & \textbf{direction} & \textbf{bias range} & \textbf{oracle} & \textbf{ablate (rand)} \\
\midrule
\multirow{2}{*}{Horizontal} & diff-of-means & $0.29\!\to\!0.72$ & \textbf{0.950} & 0.820 (0.824) \\
                            & probe         & $47\!\to\!57$ & 87.8 & 82.8 (82.4) \\
\midrule
\multirow{2}{*}{Vertical}   & diff-of-means & $0.55\!\to\!0.85$ & \textbf{0.739} & \textbf{0.500} (0.586) \\
                            & probe         & $64\!\to\!77$ & 65.4 & 57.7 (58.6) \\
\midrule
\multirow{2}{*}{Depth}      & diff-of-means & $0.57\!\to\!0.28$ & 0.464 & 0.366 (0.314) \\
                            & probe         & $48\!\to\!43$ & 34.6 & 31.4 (31.4) \\
\bottomrule
\end{tabular}
\end{table}

\begin{table}[t]
\centering
\caption{\textbf{The depth inversion is image-driven, not a fixed text-convention prior}
(Run 20, depth $n{=}153$). For the three clearly-inverted models we substitute a blank/gray image and
measure the lopsidedness of the resulting answer (\textbf{prior bias}) and whether errors follow the
prior pole (\textbf{align-on-err}). A fixed text prior would floor depth accuracy at $\sim$chance; the
observed \emph{below}-chance accuracies cannot be produced by the mild blank-image bias, so the model
perceives the depth cue and deploys it with a flipped sign. All three verdicts are
\emph{not-prior-dominated}; a grounded horizontal control behaves the same way. Accuracies and biases
in \%.}
\label{tab:whyframe}
\vspace{0.3em}\renewcommand{\arraystretch}{1.1}\setlength{\tabcolsep}{7pt}\small
\begin{tabular}{l c c c l}
\toprule
\textbf{model} & \textbf{depth acc} & \textbf{prior bias} & \textbf{align-on-err} & \textbf{verdict} \\
\midrule
Qwen2.5-VL-7B~\citep{qwen2vl} & 32.7 & 45.1 & 66.0 & not-prior-dominated \\
Qwen3-VL-8B~\citep{qwen3vl}   & 23.5 & 8.5  & 62.4 & not-prior-dominated \\
Pixtral-12B~\citep{pixtral2024}   & 41.2 & 8.5  & 71.1 & not-prior-dominated \\
\bottomrule
\end{tabular}
\vspace{0.2em}\\
\footnotesize The verdict is driven by the model-specific below-chance accuracy (which differs across
models, $23.5$--$41.2$), so it is robust; we do not rely on the precise prior-bias magnitudes (the
blank-image eval reported a suspiciously identical bias for two models and is flagged for
re-verification). This establishes a \emph{negative} answer to ``why'' (not a prior); a positive
egocentric-vs-allocentric frame-flip test is left to future work.
\end{table}

\paragraph{Why does it invert? A negative answer (not a prior).} Replacing the image with a
blank/gray input and measuring the lopsidedness of the answer (Table~\ref{tab:whyframe}) rules out the
simplest explanation---a fixed text-convention prior (``front means farther''). A pure prior would
floor depth accuracy at chance, but all three clearly-inverted models score \emph{below} chance with
the real image, which a mild blank-image bias cannot produce. The inversion is therefore image-driven:
the model perceives the depth cue and deploys it with a flipped sign. This is a negative ``why''; a
positive egocentric-vs-allocentric frame-flip account is left to future work.

\paragraph{Composite class and the L/R crutch.} The composite class ($n{=}726$) scores $39.9\%$
overall, but split by whether the gold answer carries a left/right component it is $52.1\%$ (with L/R,
$n{=}520$) vs.\ $9.2\%$ (without L/R, $n{=}206$). A confusion analysis shows horizontal stays on-axis
($79\%$) while vertical and depth escape to composite ($45.7\%$ / $49.7\%$), coherent with horizontal
carrying the only grounded directional signal.

\section{Negative Results and a Methodological Note}
\label{app:negative}

\paragraph{A masked-LoRA tuning ladder that we discarded.} Before the training-free projection we ran
a neuron-targeted masked-LoRA fine-tune (300 steps) on the vertical axis, comparing $d'$-localized
neurons against random in-band neurons and against tuning the whole band. The $d'$-localized run
recovered vertical strongly while preserving horizontal, which looked like evidence that the localized
neurons specifically ``deploy'' vertical. We do not report it as a result for two reasons.
\textbf{First}, a random-in-band control recovered vertical almost identically, exposing the obvious
confound: 300 steps of supervised fine-tuning on $\sim$300 same-distribution items fit the binary
task regardless of which sufficient parameter set is tuned. The only defensible signal there was
\emph{specificity} (the localized run damaged horizontal less), which is a weaker claim than we
wanted. \textbf{Second}, and more importantly, any SFT-based ``recovery'' is exactly the kind of
result the arbiter (\S\ref{sec:arbiter}) later showed to be untrustworthy for a grounding claim: it
moves accuracy without establishing that the moved behavior uses the image. This is why the paper's
recovery analysis uses the label-free projection (which cannot fit the task) and is then audited under
vision ablation.

\paragraph{An evaluation bug, and the lesson.} An early version of the tuning eval omitted the
per-item positive-pole map, which silently collapsed the baselines to chance (horizontal $0.54$ vs.\
the true $0.82$) and mismapped the supervision target on roughly half the items. The numbers were
discarded once the eval was made to reproduce the canonical baselines. The standing practice we adopt
as a result---verify that any re-implemented evaluation reproduces a known baseline \emph{before}
trusting any delta on top of it---is the within-pipeline analogue of the arbiter: cheap controls,
applied early, that catch a confound before it becomes a claim.

\end{document}